\documentclass[a4paper,fleqn]{cas-dc}

\usepackage[numbers]{natbib}
\usepackage[utf8]{inputenc}
\usepackage{amsmath}
\usepackage[ruled,vlined]{algorithm2e}
\usepackage[capitalise]{cleveref}
\usepackage{booktabs}
\usepackage{multirow}
\usepackage{pifont}
\usepackage{adjustbox}
\usepackage{subcaption}
\usepackage{graphicx}
\usepackage{todonotes}

\usepackage[final]{changes}
\definechangesauthor[name=del, color=red]{del}
\newcommand{\stkout}[1]{\ifmmode\text{\sout{\ensuremath{#1}}}\else\sout{#1}\fi}
\setdeletedmarkup{\stkout{#1}}

\def\tsc#1{\csdef{#1}{\textsc{\lowercase{#1}}\xspace}}
\tsc{WGM}
\tsc{QE}
\tsc{EP}
\tsc{PMS}
\tsc{BEC}
\tsc{DE}

\begin{document}
\let\WriteBookmarks\relax
\def\floatpagepagefraction{1}
\def\textpagefraction{.001}

\shorttitle{MERLIN}

\shortauthors{Nweye, Sankaranarayanan and Nagy}

\title [mode = title]{MERLIN: Multi-agent offline and transfer learning for occupant-centric energy flexible operation of grid-interactive communities using smart meter data and CityLearn}

\author[1]{Kingsley Nweye}[style=english,orcid=0000-0003-1239-5540]
\ead{nweye@utexas.edu}

\author[2]{Siva Sankaranarayanan}[style=english]
\ead{ssankaranarayanan@epri.com}

\author[1]{Zoltan Nagy}[style=english,orcid=0000-0002-6014-3228]
\ead{nagy@utexas.edu}
\cormark[1]
\cortext[cor1]{Corresponding author}

\affiliation[1]{organization={Intelligent Environments Laboratory\\Department of Civil, Architectural and Environmental Engineering\\
  The University of Texas at Austin},
    addressline={301 E. Dean Keeton St., ECJ 4.200}, 
    city={Austin},
    postcode={78712-1700}, 
    state={Texas},
    country={USA}
}

\affiliation[2]{organization={Electric Power Research Institute},
    city={Palo Alto},
    state={California},
    country={USA}
}

\begin{abstract}
The decarbonization of buildings presents new challenges for the reliability of the electrical grid as a result of the intermittency of renewable energy sources and increase in grid load brought about by end-use electrification. To restore reliability, grid-interactive efficient buildings can provide flexibility services to the grid through demand response. Residential demand response programs are hindered by the need for manual intervention by customers. To maximize the energy flexibility potential of residential buildings, an advanced control architecture is needed. Reinforcement learning is well-suited for the control of flexible resources as it is able to adapt to unique building characteristics compared to expert systems. Yet, factors hindering the adoption of RL in real-world applications include its large data requirements for training, control security and generalizability. Here we address these challenges by proposing the MERLIN framework and using a digital twin of a real-world 17-building grid-interactive residential community in CityLearn. We show that 1) independent RL-controllers for batteries improve building and district level KPIs compared to a reference RBC by tailoring their policies to individual buildings, 2) despite unique occupant behaviours, transferring the RL policy of any one of the buildings to other buildings provides comparable performance while reducing the cost of training, 3) training RL-controllers on limited temporal data that does not capture full seasonality in occupant behaviour has little effect on performance. Although, the zero-net-energy (ZNE) condition of the buildings could be maintained or worsened as a result of controlled batteries, KPIs that are typically improved by ZNE condition (electricity price and carbon emissions) are further improved when the batteries are managed by an advanced controller.
\end{abstract}




\maketitle
\section{Introduction} \label{sec:introduction}
The residential building stock in the United States is responsible for 21\% of energy consumption \cite{eia-mer-00352211} and 20\% of greenhouse gas (GHG) emissions \cite{goldstein2020carbon}. Electrification of end-uses, as well as decarbonizing the electrical grid through renewable energy sources such as solar and wind, constitutes the pathway to zero-emission buildings \cite{Lanham2018}. However, the intermittency of renewable energy sources introduces additional challenges of grid instability for decarbonization due to mismatch between electricity generation and demand \cite{YEKINISUBERU2014499} and  could reduce the economic value of such sources \cite{doi:10.1086/686733}.


Buildings can provide flexibility services to the grid such as energy efficiency, load shifting, peak shaving, and load modulation, and have been been referred to as grid-interactive efficient buildings (GEBs) in the literature \cite{osti_1508212}. These services are activated through participation of the electricity market customers of various sectors in demand-response programs that may be incentive-based (direct) e.g., direct-load control (DLC) or price-based (indirect) e.g., time-of-use (TOU) \cite{NIKZAD201483}. While price-based programs do not suffer from the privacy issues stemming from direct control in incentive-based programs they require customer intervention and comprehension of electricity rate structures for their success , thus, automated demand-response participation may help in its widespread adoption.

Demand response has existed amongst commercial and industrial customers as they operate large machinery and have high energy use intensities (EUIs) that could provide ample grid flexibility in times of supply deficit. On the other hand, demand response by residential customers is relatively new and is provided through the control of flexible energy resources e.g., cooling and heating systems whose energy consumption may be curtailed through the adjustment of thermostat set-points, electric vehicles (EV) that can alter their charge rate or provide power to the grid in Vehicle-to-Grid (V2G) scenarios, passive storage systems e.g., thermal mass that can provide pre-cooling and pre-heating services or active storage systems such as thermal storage and battery systems that act as temporal energy buffers. Residential buildings may also provide grid services through self-generation using solar photovoltaics (PVs). Demand response programs in residential buildings may require manual intervention from the consumer upon signaling from the grid, which may bring about lack of participation in such programs \cite{HAIDER2016166}. The diffusion of interconnected and smart energy appliances and systems in residential buildings provides opportunities for intelligent building control via a home energy management system (HEMS), providing efficient energy use, automated grid services and comfort.

Furthermore, residential buildings participating independently in a demand response program may not provide the best performance at the aggregated transformer or grid level. In fact, unwanted effects may occur such as shifting peak demand to an earlier time rather then reducing the peak \cite{Gelazanskas2014DemandDirection}. To improve performance and maximize value, aggregators, which are recent actors in the electricity market, can act as middlemen between grid operators and end-users (consumers and prosumers) by coordinating and selling energy flexibility between both parties e.g., thermostat set-points or electricity storage in the electricity market \cite{BURGER2017395}. Customers are then rewarded for their traded flexibility. The aggregation of buildings act as virtual power plants (VPP).


Advanced control algorithms such as model predictive control (MPC)~\cite{Drgona2020} and deep reinforcement learning (RL)~\cite{Wang2020} have been proposed for a variety of building control applications. While both  methods have their disadvantages e.g., MPC requiring a model while RL being data intensive, notable applications and results have been presented in the past several years for MPC \cite{privara2011model, karlsson2011application, yuan2006multiple} and RL \cite{Yang2015a, o2010residential, liu2006experimental, mozer1998neural, costanzo2016experimental, ruelens2016reinforcement}. In addition, hybrid methods, e.g., based on physics-constrained neural networks for models~\cite{Drgona2021}, or that merge RL and MPC approaches\cite{Arroyo2022, Chen2020a},  have recently emerged.

In contrast to MPC, RL is an adaptive and potentially model-free control algorithm that can take advantage of both real-time and historical data. RL is an agent-based machine learning algorithm in which the agent learns optimal actions via interaction with its environment~\cite{RichardS.SuttonandAndrewG.Barto2018a, Nagy2018}. In contrast to supervised learning, the agent does not receive large amounts of labelled data to learn from. In contrast to unsupervised learning, the agent receives delayed feedback from the environment. These characteristics of RL situates it as a preferable control algorithm for use by aggregators to maximize the value from distributed energy resources that are provided by their customers. 

While the advancements in RL have been noteworthy in recent years, demonstrating human-level or better performance in particular in video games~\cite{mnih15, Silver2017, Vinyals2019}, real world applications, especially for complex systems, have been rare. In their review, \citeauthor{wang2020reinforcement} identified three barriers hindering the adoption of RL in building controls to include 1) time-consuming and data-intensive training requirements, 2) control security and robustness and 3) generalizability of RL controllers \cite{wang2020reinforcement}. RL suffers from the sample efficiency problem where large amounts of historical observations are need to train deep neural networks in order to achieve satisfactory convergence and performance. Our previous work has investigated the use of expert control systems such as rule-based controllers (RBCs) to provide training data for offline learning \cite{NWEYE2022100202} or synthetic data to augment limited available training data \cite{10.1145/3563357.3564080}. The control security barrier refers to ensuring that the actions taken by the controller are not catastrophic in nature where they may result it occupant discomfort or wastage of resources. Back-up controllers that override the RL controller actions are typically employed to provide control security and robustness. Alternatively, pre-training on expert systems e.g. RBC could be used to teach the control agent typical decision trajectories \cite{wang2020reinforcement}. The third barrier, generalizability of RL controllers, stems from the uniqueness of buildings thus requiring custom RL controllers for each building. Transfer learning is an area of machine learning that is interested in leveraging the knowledge of a source environment's domain and learning task to improve learning of a target environment's predictive function for its learning task where the source and target environments' domain or learning task are different \cite{pinto2022transfer}. In the context of system control in grid interactive communities, the source and target environments could refer to buildings in the communities, the domain could refers to variables in the building e.g., weather variables, battery state-of-charge (SOC) and other observable states, and their probability distributions. Whereas, the learning task could refer to the control action and the function or policy that maps the observations to the appropriate control actions. With transfer learning, one source building or multiple source buildings in a grid-interactive community can be used to train and develop generalizable RL control policies that could be deployed live in other target buildings in the community thus, reducing the cost of acquiring domain and learning task samples in the target buildings. We refer the readers to \cite{pinto2022transfer} for a comprehensive review of transfer learning in smart buildings.

In this work, we address the three barriers highlighted in \cite{wang2020reinforcement} that hinder the real-world adoption of RL in building controls. We propose MERLIN, a framework for \textit{Multi-agent offline and transfER Learning for occupant-centric energy flexible operation of grid-INteractive communities using smart meter data and CityLearn} that can be used by aggregators to easily train and deploy control policies for customers' DERs. Our implementation example utilizes smart meter data from a real world grid-interactive community of 17 zero-net energy (ZNE) single-family homes and shows the application of batteries controlled by an RL policy in the CityLearn environment for \textit{continuous} demand response \cite{citylearnArxiv} to minimize customer electricity consumption, bill and emissions and provide provide grid flexibility. With MERLIN and a digital twin of the community in CityLearn, we show that 1) comfort can be maintained and unique occupant behaviour can be catered to by a system-constraint aware RL control architecture while maximizing the flexibility value of controlled batteries 2) with only five months of data for offline training, comparable performance to a full year of data that captures the seasonality in occupant behaviour and weather can be achieved and 3) satisfactory building and district-level objectives can be achieved when a source building's RL control policy is transferred to other target buildings.

Thus, we structure this work as follows: in \cref{sec:framework} we provide an overview of the MERLIN framework pertaining to its data requirements, simulation environment and performance evaluation criteria. In \cref{sec:implementation} we describe the datasets, environment and agent set up that are used in a our case-study demonstration of MERLIN as well as the performance cost functions that we use to evaluate our results in \cref{sec:results}. We then contextualize or presented results in \cref{sec:discussion} and conclude in \cref{sec:conclusion}.









\section{MERLIN Framework} 
\label{sec:framework}
\begin{figure*}[tbh]
    \centering
    \includegraphics[width=\textwidth]{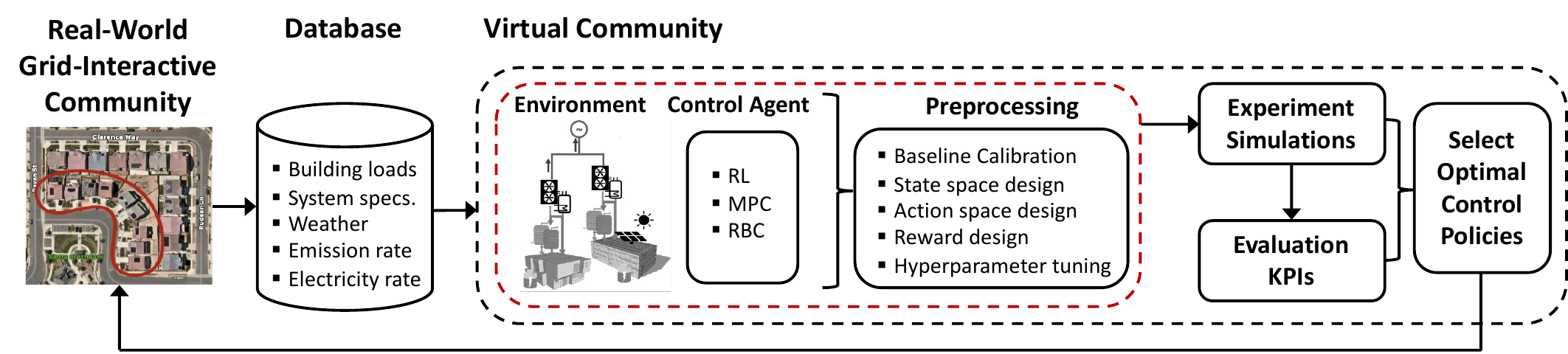}
    \caption{MERLIN framework.}
    \label{fig:methodology}
\end{figure*}

MERLIN (\cref{fig:methodology}) is designed to provide a framework for the training, evaluation and deployment of DER control policies by an aggregator or similar operator in grid-interactive communities. These framework objectives are achieved through simulation of the grid-interactive community in a control environment with the goal of optimizing a set of cost functions or key performance indicators (KPIs).

The MERLIN framework begins with the collection of data that are typically required for energy model creation and calibration including buildings' energy loads, distributed energy resources and other building energy system specifications, weather, electricity rate and and carbon emissions. Building energy loads can be obtained through simulation of energy models in energy simulation software e.g., EnergyPlus or real-world measurements in smart meters. Building energy system specifications such as thermal storage and battery capacities are either sourced directly from the buildings or estimated. Weather data, carbon emission rate and electricity rate are other data that may be used as states or for KPI evaluation in the simulation environment.

The digital twin that is then created using the collected data is a simulation environment under the manipulation of an expert control agent e.g. rule-based controller (RBC) or advanced control agent e.g reinforcement learning controller (RLC) and model predictive controller (MPC). Emulators that are designed for control algorithm benchmarking such as CityLearn \cite{citylearnArxiv}, BOPTEST \cite{doi:10.1080/19401493.2021.1986574}, ACTB \cite{Marzullo2022} and Energym \cite{scharnhorst2021energym} provide the simulation environment and the environment contain models of buildings and their energy systems such as heat pumps, thermal storage systems, batteries, electric heaters, PVs, and other DERs that are controlled to manage building-load satisfaction and provide energy flexibility. In this work, we make use of the CityLearn environment as its advantage over other emulators is that it is purposely built for control applications that work with smart meter data, rather than as a co-simulation environment with design software such as EnergyPlus. This makes CityLearn a powerful alternative for studying data-driven control systems in real-world applications.

The simulation environment and control agent are then validated before carrying out benchmarking tasks. Such validation could include calibration of the environment with pre-measured data, action and state (observation) space design for the control agent, reward function design and agent hyper-parameter grid search. 

Once validation is completed, experiments are carried out to determine what control policy has the best performance. A best-performing control policy for a building is one that yields the best outcome for a set of KPIs, which quantify energy flexibility at the building and grid (district) levels.

Finally, the best-performing control policies are deployed in the real-world grid-interactive community in the respective building whose data they were trained on or used as source environment to be transferred and leveraged in other target buildings.

\section{MERLIN Implementation} \label{sec:implementation}
\begin{figure}
    \begin{subfigure}[]{.345\columnwidth}
        \centering
        \includegraphics[width=\textwidth]{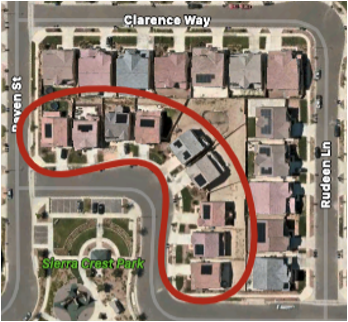}
        \caption{Lot group 1.}
    \end{subfigure}\hfill
    \begin{subfigure}[]{.645\columnwidth}
        \centering
        \includegraphics[width=\textwidth]{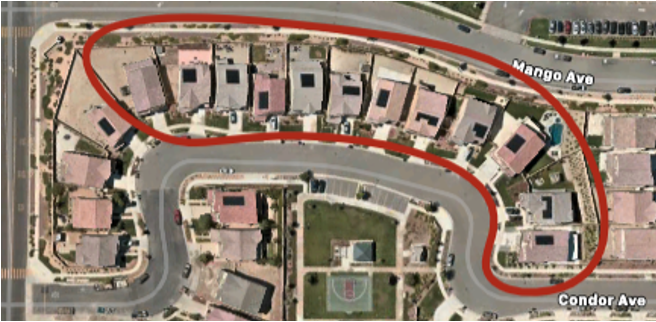}
        \caption{Lot group 2.}
    \end{subfigure}
    \begin{subfigure}[]{\columnwidth}
        \centering
        \includegraphics[width=\textwidth]{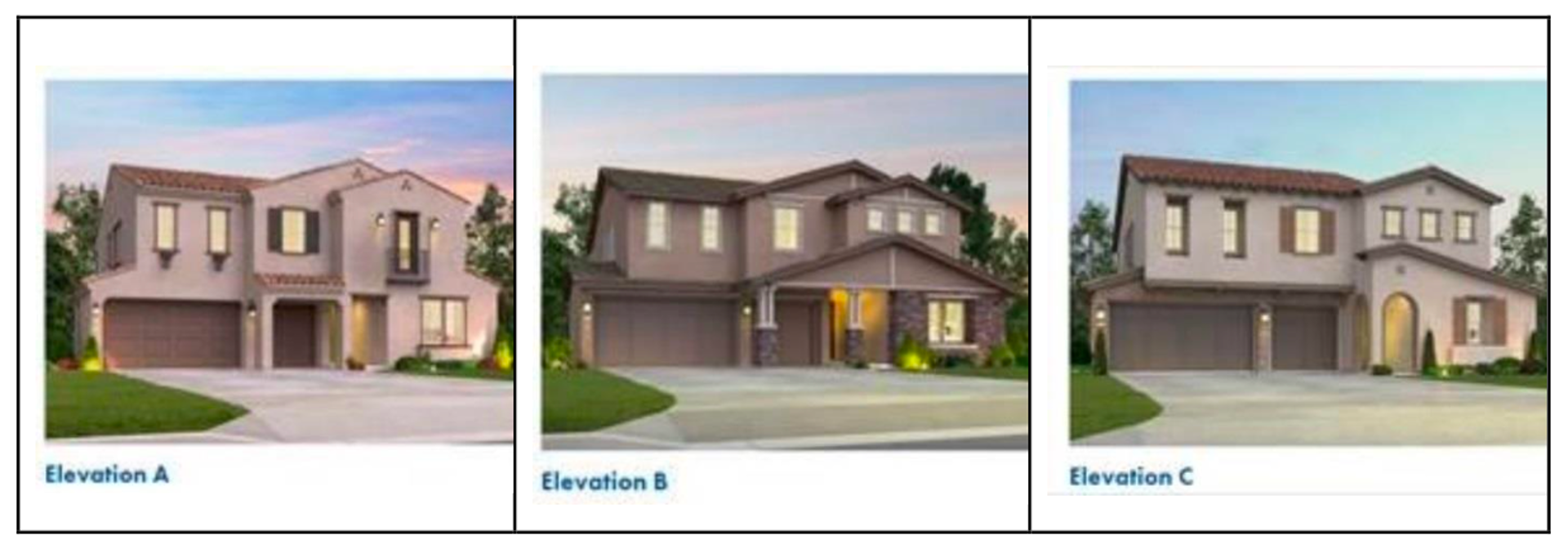}
        \caption{Building elevation varieties.}
    \end{subfigure}
    \caption{17 case-study ZNE buildings in the Sierra Crest Zero Net Energy community in Fontana, California.}
    \label{fig:fontana_location}
\end{figure}

We implement MERLIN using data from 17 ZNE single-family homes in the Sierra Crest Zero Net Energy community in Fontana, California, which is pictured in \cref{fig:fontana_location} \cite{epriFontana2017}. The buildings were studied for grid integration of zero net energy communities as part of the California Solar Initiative program specifically exploring the impact of high penetration PV generation and on-site electricity storage \cite{narayanamurthy2016grid}. 

\begin{figure}
    \centering
    \includegraphics[width=\columnwidth]{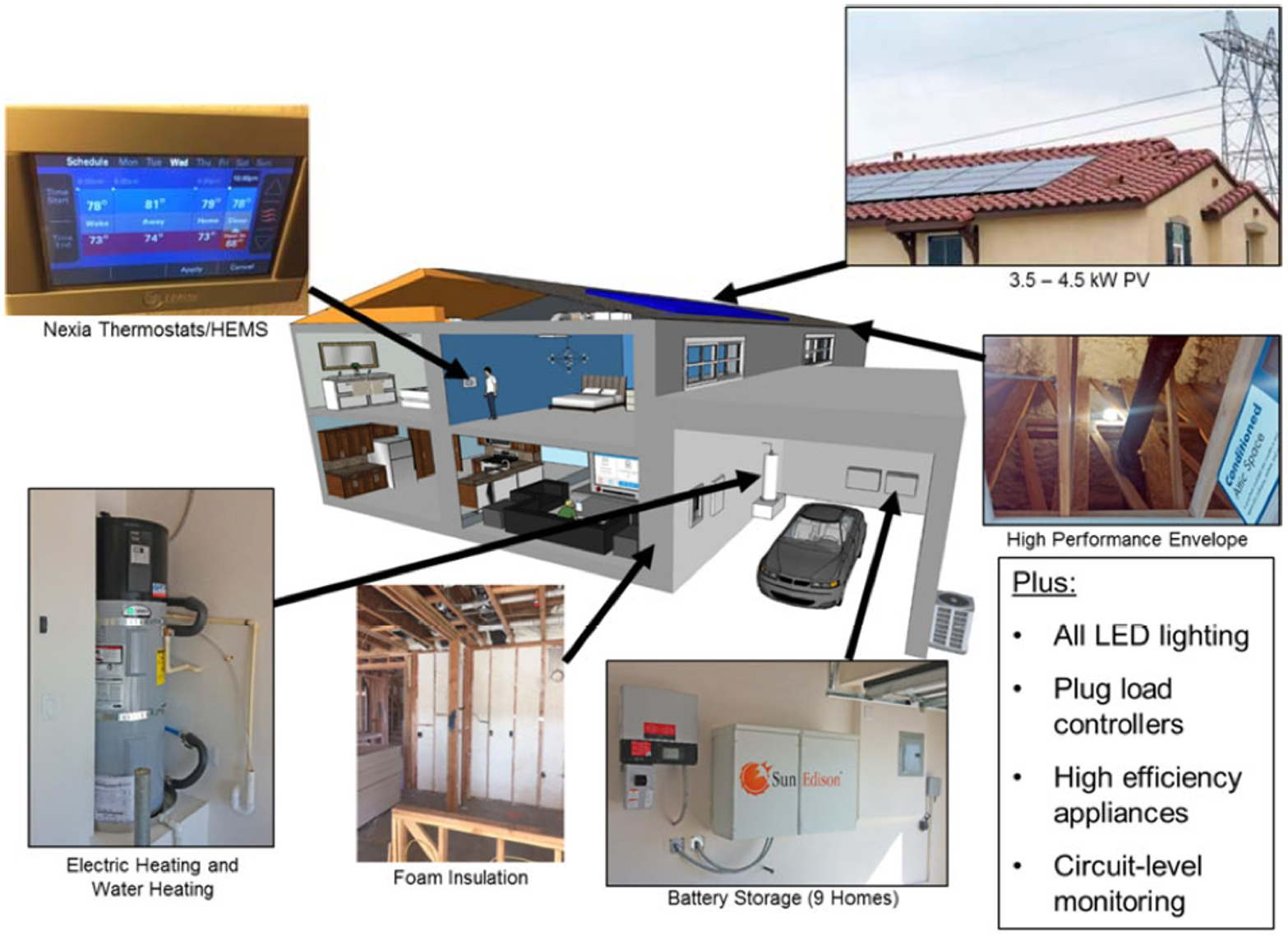}
    \caption{Envelope and system characteristics of case-study ZNE building.}
    \label{fig:fontana_building_characteristics}
\end{figure}

Each building is a single-family archetype that was constructed in the mid to late 2010s and has a gross floor area between 177 m\textsuperscript{2} and 269 m\textsuperscript{2}. \cref{fig:fontana_building_characteristics} shows the envelope and system characteristics of the buildings in the Sierra Crest Zero Net Energy community. The building envelope is made up of high-performance materials for high energy efficiency and the buildings are equipped with high-efficiency appliances, electric heating and water heating systems. The buildings also have circuit-level monitoring that provide high resolution energy use time series data as well as and home energy management systems (HEMS) for occupant control. In the as-built community, eight of the 17 buildings are equipped with 6.4 kWh capacity batteries that have a 5 kW power rating, 90\% round-trip efficiency, and 75\% depth-of-discharge. The installed PV capacity is 4 kW or 5 kW for homes with or without batteries, respectively.

We create a digital twin of the buildings under study to evaluate three deployment strategies that are distinguished by the  quantity of data made available to the RL agents for training and scope of target buildings for deployment.


We describe the datasets used for the digital twin in \cref{sec:implementation-database} and provide information about preprocessing steps, simulation environment and control-agent in \cref{sec:implementation-environment,sec:implementation-reference_rbc,sec:implementation-agent_design}. The deployment strategies are described in detail in \cref{sec:implementation-deployment_strategies-strategy_1,sec:implementation-deployment_strategies-strategy_2,sec:implementation-deployment_strategies-strategy_3} while the evaluation KPIs are defined mathematically in \cref{sec:implementation-kpi}.

\subsection{Datasets} \label{sec:implementation-database}
We utilize a novel one-year time series dataset from the 17 case-study ZNE buildings that covers the August 1, 2016 to July 31, 2017 period. This time series dataset was used in the \textit{NeurIPS 2022 -- CityLearn Challenge} \cite{AICrowdCityLearnChallenge2022}, the third edition of the annual \textit{The CityLearn Challenge} \cite{Vazquez-Canteli2020a, Nagy2021}. The net building demand is measured at a 1-minute resolution and is an aggregate of plug demand, space heating and cooling demand, domestic hot water demand, PV supply power and battery charging/discharging demand (where applicable). In this work we down-sample the time-series data to hourly resolution to speed up the simulation process and convert the power, $P$ (kW) measurements to energy, $E$ (kWh) unit then re-sample to hourly resolution using \cref{eqn:power_to_energy_conversion} where $P_m$ is the power at minute $m$ and $E_h$ is the energy consumption at hour $h$. We then calculate non-shiftable end-use load that must be satisfied irrespective of solar generation and battery control policy using \cref{eqn:non_shiftable_load}.

\begin{equation}
    E_h = \sum_m^{60}{\frac{P_m}{60}}
    \label{eqn:power_to_energy_conversion}
\end{equation}

\begin{equation}
    E_h^\textrm{non-shiftable} = E_h^\textrm{main} - \left(E_h^\textrm{battery} + E_h^\textrm{PV}\right) 
    \label{eqn:non_shiftable_load}
\end{equation}

Other supplemental data collected for the simulation environment include TMY3 weather data from the Los Angeles International Airport weather station and carbon intensity (kg\textsubscript{CO\textsubscript{2}e}/kWh) time-series. The electricity rate-plan we utilize is that of the community's utility provider, Southern California Edison. We adopt their \textit{TOU-D-PRIME} rate plan summarized in \cref{tab:elecricity_rate}, and designed for customers with residential batteries \cite{sCEDTOU} where electricity is cheapest in the early morning and late night and cheaper during off-peak months of October-May. Meanwhile, electricity is cheaper on weekends for peak hours of 4 PM-9 PM.

\begin{table}[h]
    \centering
    \caption{Time-Of-Use rate plan (\$/kWh) used as electricity rate in simulation environment.}
    \label{tab:elecricity_rate}
    \begin{tabular}{@{}lrrrr@{}}
        \toprule
        \multicolumn{1}{c}{\multirow{2}{*}{Time}} & \multicolumn{2}{c}{June-September} & \multicolumn{2}{c}{October-May} \\ 
        \cmidrule(l){2-5} 
        \multicolumn{1}{c}{} & \multicolumn{1}{c}{Weekday} & \multicolumn{1}{c}{Weekend} & \multicolumn{1}{c}{Weekday} & \multicolumn{1}{c}{Weekend} \\ 
        \midrule
        8 AM-4 PM & 0.21 & 0.21 & 0.20 & 0.20 \\
        4 PM-9 PM & 0.54 & 0.40 & 0.50 & 0.50 \\
        9 PM-8 AM & 0.21 & 0.21 & 0.20 & 0.20 \\ 
        \bottomrule
    \end{tabular}
\end{table}

Using the measured building demand data, battery and PV system specifications, and supplemental data, an equivalent digital twin of the 17 buildings is generated in CityLearn (see \cref{sec:implementation-environment}).

\subsection{CityLearn} \label{sec:implementation-environment}
The digital twin is created in CityLearn, an OpenAI Gym environment that was created for the benchmarking of RL algorithms for demand response studies \cite{citylearnArxiv}. CityLearn has been used extensively as a reference environment to demonstrate incentive-based DR \cite{Deltetto2021}, collaborative DR \cite{Glatt2021}, coordinated energy management \cite{Pinto2021, Kathirgamanathan2020, pinto2021data}, benchmarking DR algorithms \cite{Dhamankar2020, Qin2021, pinto2022enhancing} and voltage regulation \cite{PIGOTT2022108521}.

CityLearn is designed to provide \textit{continuous} demand response without the need for a demand response signal from the grid (though it is possible to also include that) by intelligently leveraging active storage systems for load shifting. It has models of buildings, electric heaters, heat pumps, thermal storage, batteries and PVs. In our application of CityLearn, the environment includes energy system models of 17 buildings, where each building is equipped with a battery and PV system that have the same specifications as the real-world community buildings. The batteries are used to store energy that may be eventually discharged to satisfy net positive building loads, and are charged by the grid and/or PV system, if generated power is available in surplus. Note that while the as-built community did not equip each building with a battery, our implementation does so.

The control agents manage the battery's SOC by prescribing how much energy to store or release at any given time step. CityLearn guarantees that loads are satisfied irrespective of the agent's actions since the building load is known a-priori through the measured or simulated load data. An internal backup controller also ensure that system constraints such as occupant comfort and base loads are never violated by ensuring that building load is first satisfied before charging the battery and that the battery never discharges more energy than needed to satisfy the building load.


\subsection{Reference Rule-Based Controller Design} \label{sec:implementation-reference_rbc}
We utilize a rule-based controller (RBC) to provide a reference expert system control policy for performance comparison with the RL alternative. The buildings' batteries in the Sierra Crest Zero Net Energy community operate on one of three control strategies \cite{narayanamurthy2016grid}. The \textit{Self-Consumption} strategy is to charge the battery during net export and discharge at other times. The \textit{Time-of-Use Peak Reduction} strategy is to charge between 9:00 AM and 12:00 PM then discharge at 6:00 PM at constant 2 kW rate. Lastly, the \textit{Time-of-Use Rate Optimization} strategy is to charge from 6:30 PM at 4.5 kW maximum rate then discharge from 12:00 PM. All strategies maintain 25\% SOC and differ in the time of and conditions for charge and discharge. 

However, since the buildings are anonymized, it is unknown which of the three control strategies is used in each building. Also, the CityLearn environment allows for one control architecture to be applied to all buildings. Thus, to determine what control strategy to use as a reference, we simulate the three battery control strategies in a subset of the 17 buildings, and the strategy that minimizes the aggregated community error between the simulated and measured battery electricity consumption is used as the reference RBC for the remainder of this work. 

There are eight buildings equipped with a battery in the as-built community and their hourly battery electricity consumption profiles, calculated from smart meter-measured battery demand using \cref{eqn:power_to_energy_conversion}, are shown in \cref{fig:measured_battery_electricity_consumption}. Although batteries are installed in eight buildings, only the batteries in buildings 6/8 are operational while those in 1 and 5 are not in use. Hence, we use only data from buildings with operational batteries i.e. buildings 2, 3, 6, 7, 8 and 9, for the reference RBC validation. Furthermore, there are instances in most buildings when the battery consumption is $\approx 0.0\textrm{kWh}$ throughout a 24-hour daily period. We exclude such days from the validation such that any day during which the absolute sum of calculated battery electricity consumption is $\le 1.0\textrm{kWh}$ is not used to calculate the simulation error.

We refer the reader to \cref{fig:rbc_validation} for a summary of the results from the reference RBC policy determination. We find that the \textit{Time-of-Use Peak Reduction} control strategy i.e., to charge between 9:00 AM and 12:00 PM then discharge at 6:00 PM at constant 2 kW rate, minimizes the error between simulated and measured battery electricity consumption. Thus, we use the \textit{Time-of-Use Peak Reduction} strategy as our reference RBC.

\subsection{Reinforcement Learning Agent Design} \label{sec:implementation-agent_design}
Each building's battery is controlled by an independent Soft actor-critic (SAC) agent. SAC is a model-free off-policy RL algorithm \cite{Haarnoja2018}. As an off-policy method, SAC can reuse experience and learn from fewer samples. SAC is based on three key elements: an actor-critic architecture, off-policy updates, and entropy maximization for efficient exploration and stable training. SAC learns three different functions: the actor (policy), the critic (soft Q-function), and the value function $V$. For more details about SAC, we refer the reader to \cite{Haarnoja2018SoftAA}. 

Typically, RL algorithms are initialized with random samples for the weights in the neural networks, which are then determined over many episodes. However, as we have shown in our previous work \cite{Vazquez-Canteli2020, NWEYE2022100202}, we can jump-start the RL algorithm offline with an available RBC policy and achieve the same performance, while greatly reducing the learning time. This is a realistic approach for building energy system management as such systems are equipped with an expert control policy by default e.g., RBC. Therefore, in our implementation here, we modify the SAC algorithm by utilizing the reference RBC policy (\cref{sec:implementation-reference_rbc}) for action selection during the initial 3,671 exploration time steps (five months) of offline training. In addition to reducing training time, this also improves the robustness of the RL agents, as they do not take too random actions.
 
We further describe the RL agent hyperparameters, observation space, action space and reward function in \cref{sec:implementation-hyperparameter_tuning,sec:implementation-observation_and_action_space_design,sec:implementation:reward_design}.

\subsubsection{Hyperparameter Tuning} \label{sec:implementation-hyperparameter_tuning}
The performance of RL algorithms in a control environment is sensitive to the choice of hyperparameter values thus, the need for hyperparameter tuning. The hyperparameters in \cref{tab:agent_fixed_hyperparameter} are fixed while we use a grid search approach to determine the best-performing values for sensitive hyperparameters shown in \cref{tab:agent_hyperparameter_search_grid}. The decay rate, $\tau$, sets the magnitude at which the SAC target network is updated. The discount factor, $\gamma, \in [0, 1]$ is used to balance between an agent that considers only immediate rewards ($\gamma = 0$) and one that aims to optimize future rewards ($\gamma = 1$). The learning rate, $\alpha, \in [0, 1]$ refers to the rate at which Q-values are updated i.e., old knowledge replacing new knowledge. Increasing $\alpha$ leads to more loss of previous knowledge. The temperature, $T, \in [0, 1]$ provides a balance between completely exploitative ($T = 0$) and completely exploratory ($T = 1$) action selection. Each combination of $\tau$, $\gamma$, $\alpha$ and $T$ is evaluated three times with distinct random state seeds for a CityLearn district that consists of the same six buildings we use for reference RBC validation; and the combination that maximizes the reward sum is selected.

\begin{table}[]
\centering
\caption{SAC agent fixed hyperparameter.}
\label{tab:agent_fixed_hyperparameter}
\begin{tabular}{lr}
\toprule
Variable & Value \\ 
\midrule
Batch size & 256 \\
Neural network hidden layer count & 2 \\
Neural network hidden layer size & 256 \\
Replay buffer capacity & 100,000 \\
Time steps & 8,760 (1 year) \\
Training episodes & 10 \\
\bottomrule
\end{tabular}
\end{table}

\begin{table}[]
\centering
\caption{SAC agent hyperparameter search grid.}
\label{tab:agent_hyperparameter_search_grid}
\begin{tabular}{ll}
\toprule
Variable & Value grid \\ 
\midrule
Decay rate ($\tau$) & [0.0005, 0.005, 0.05] \\
Discount factor ($\gamma$) & [0.90, 0.95, 0.99] \\
Learning rate ($\alpha$) & [0.00005, 0.0005, 0.005] \\
Temperature ($T$) & [0.2, 0.5, 0.8] \\
\bottomrule
\end{tabular}
\end{table}

We refer the reader to \cref{fig:last_episode_hyperparameter_combination_reward} for a summary of the results from the SAC agents hyperparameter grid search. In \cref{tab:hyperparameter_grid_search_selection}, we show the values of $\tau$, $gamma$, $\alpha$ and $T$ that maximize the cumulative reward sum for each building's SAC agent. The hyperparameter values used in the SAC agents for the remainder of this work are the mode values amongst the buildings: $\tau=0.05$, $\gamma=0.9$, $\alpha=0.005$ and $T=0.2$.

\begin{table}[h]
    \centering
    \caption{Selected SAC agent hyperparameters.}
    \label{tab:hyperparameter_grid_search_selection}
    \begin{tabular}{lrrrrrrr}
        \toprule
        & $\tau$ & $\gamma$ & $\alpha$ & $T$ & $\sum_{h=0}^{8759}{r}$ \\
        \midrule
        Building 2 & 0.05 & 0.9 & 0.005 & 0.2 & -1746.8 \\
        Building 3 & 0.0005 & 0.9 & 0.005 & 0.2 & -1384.1 \\
        Building 6 & 0.05 & 0.9 & 0.005 & 0.8 & -2131.9 \\
        Building 7 & 0.05 & 0.9 & 0.005 & 0.5 & -1680.5 \\
        Building 8 & 0.05 & 0.9 & 0.005 & 0.5 & -1644.9 \\
        Building 9 & 0.05 & 0.9 & 0.005 & 0.2 & -1498.9 \\
        \textbf{Mode} & \textbf{0.05} & \textbf{0.9} & \textbf{0.005} & \textbf{0.2} & - \\
        \bottomrule
    \end{tabular}
\end{table}

\subsubsection{Observation and Action Space Design} \label{sec:implementation-observation_and_action_space_design}
The agents' observation space is made up of community level temporal and weather states and, building-specific states that are listed in \cref{tab:observation_and_action_space_design}. The two temporal observations encode the time-dependent nature of the environment and are predefined in all time-series data. The weather observations include the direct solar irradiance and forecasts, which come from the supplemental weather data. Non-shiftable load is the total building load before adding solar generation and battery load. Net-electricity consumption is the sum of non-shiftable load, solar generation and battery load. The battery SOC is the ratio of stored energy to the initial battery capacity and is calculated during simulation. Carbon emission and net electricity price encode the environmental and financial costs of using the grid to satisfying building loads. The observations are transformed to aid the learning process by applying cyclical transformation, one-hot encoding or min-max normalization on periodic, discrete or continuous observations respectively.

\begin{table}[h]
    \centering
    \caption{Observation space.}
    \label{tab:observation_and_action_space_design}
    \begin{tabular}{lll}
        \toprule
        Observation & Unit & Transformation \\
        \midrule
        \textbf{Temporal} & \\
        Day   & - & One-hot  \\
        Hour  & - & Cyclical \\
        \textbf{Weather} & \\
        Dir. solar irradiance & W/m\textsuperscript{2} & Min-max norm. \\
        Dir. solar irradiance (+6 hr.) & W/m\textsuperscript{2} & Min-max norm. \\
        Dir. solar irradiance (+12 hr.) & W/m\textsuperscript{2} & Min-max norm. \\
        Dir. solar irradiance (+24 hr.) & W/m\textsuperscript{2} & Min-max norm. \\
        \textbf{Building} & \\
        Solar generation & kWh & Min-max norm. \\
        Net-electricity consumption & kWh & Min-max norm. \\
        Non-shiftable load & kWh & Min-max norm. \\
        Battery SOC & - & Min-max norm. \\
        Carbon emissions & kg\textsubscript{CO\textsubscript{2}} & Min-max norm.\\
        Net electricity price & \$ & Min-max norm. \\
        Net electricity price (+6 hr.) & \$ & Min-max norm. \\
        Net electricity price (+12 hr.) & \$ & Min-max norm. \\
        \bottomrule
    \end{tabular}
\end{table}

The action space is 1-dimensional and is the fraction of the battery capacity to be charged or discharged. Its value is bounded between -1 and 1 where positive and negative values are charge and discharge control actions respectively.

\subsubsection{Reward Design} \label{sec:implementation:reward_design}
The reward function is designed to minimize electricity price, $C$ (see \cref{eqn:kpi-electricity_price}) and carbon emissions from grid electricity supply, $G$ (see \cref{eqn:kpi-carbon_emissions}). The reward promotes net-zero energy use by penalizing grid load satisfaction when SOC\textsubscript{battery}, is non-zero as well as penalizing net export when the battery is not fully charged through the penalty term, $p$ defined in \cref{eqn:reward_multiplier}.

$C$ and $G$ are weighted using $w_1$ and $w_2$ respectively and have exponent terms, $e_1$ and $e_2$. We use a grid search approach to determine the values of $w_1$, $w_2$, $e_1$ and $e_2$. Each combination of $w_1$, $w_2$ $e_1$ and $e_2$ is evaluated three times with distinct random state seeds for a CityLearn district that consists of the same six buildings we use for reference RBC validation; and the combination that minimizes the district-level electricity price (\cref{eqn:kpi-electricity_price}) and carbon emissions (\cref{eqn:kpi-carbon_emissions}) KPIs, as well as their average, is selected.

\begin{equation}
    r = p \times \Big|w_1 C^{e_1} + w_2 G^{e_2}\Big|
    \label{eqn:reward}
\end{equation}

\begin{equation}
    p = -\left(1 + \textrm{sign}(C) \times \textrm{SOC\textsubscript{battery}}\right)
    \label{eqn:reward_multiplier}
\end{equation}

We refer the reader to \cref{fig:reward_design_last_episode_district_price_and_emission_scores} for a summary of the results from the reward parameter grid search. The parameters used in the reward function for the remainder of this work are those which minimize both $G$ and $\overline{C+G}$ as reported in \cref{tab:reward_paramters}.

\begin{table}[h]
    \centering
    \caption{Selected reward parameters, $e_1$, $e_2$, $w_1$ and $w_2$ that minimize average building electricity price, $C$, carbon emissions, $G$ and their combined average, $\overline{C+G}$.}
    \label{tab:reward_paramters}
    \begin{tabular}{lllllll}
        \toprule
        $e_1$ & $e_2$ & $w_1$ & $w_2$ & $C$ & $G$ & $\overline{C+G}$ \\
        \midrule
        1 & 1 & 1.0 & 0.0 & 0.778 & 0.865 & 0.821 \\
        \bottomrule
    \end{tabular}
\end{table}

\subsection{Deployment Strategies} \label{sec:implementation-deployment_strategies}

\subsubsection{Deployment Strategy 1 (DS.1)} \label{sec:implementation-deployment_strategies-strategy_1}
\begin{figure}
    \begin{subfigure}[]{\columnwidth}
        \centering
        \includegraphics[width=\columnwidth]{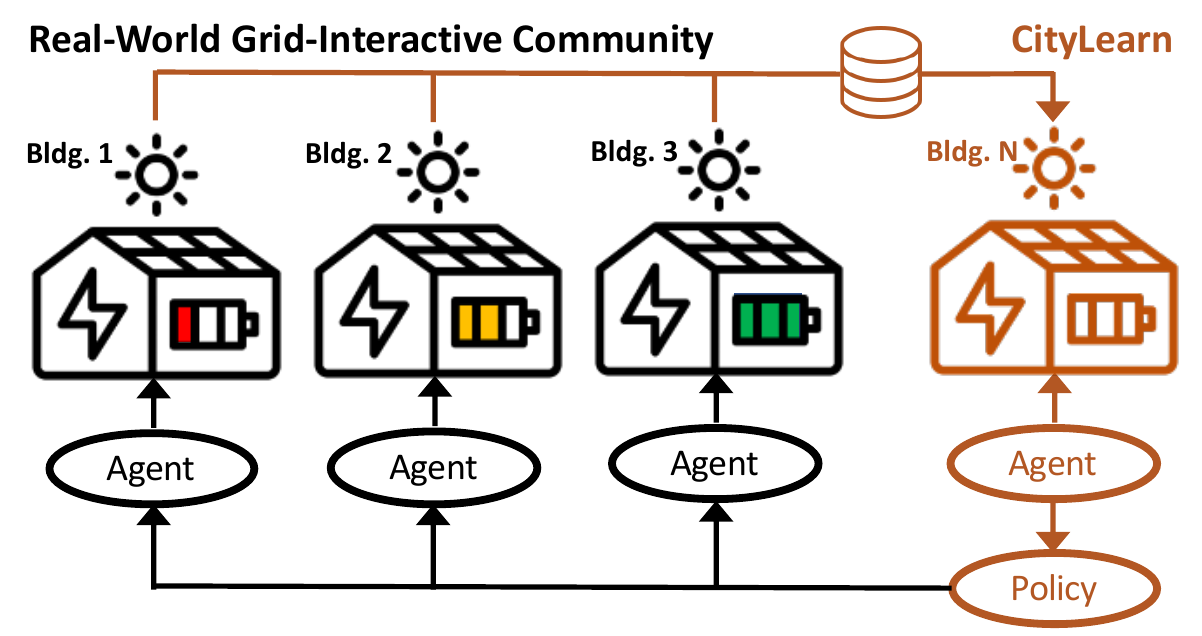}
        \caption{Deployment Strategy 1 -- all buildings share at least one year of data and a unique policy is created for each building N.}
        \label{fig:deployment_strategy_1}
    \end{subfigure}\hfill
    \begin{subfigure}[]{\columnwidth}
        \centering
        \includegraphics[width=\columnwidth]{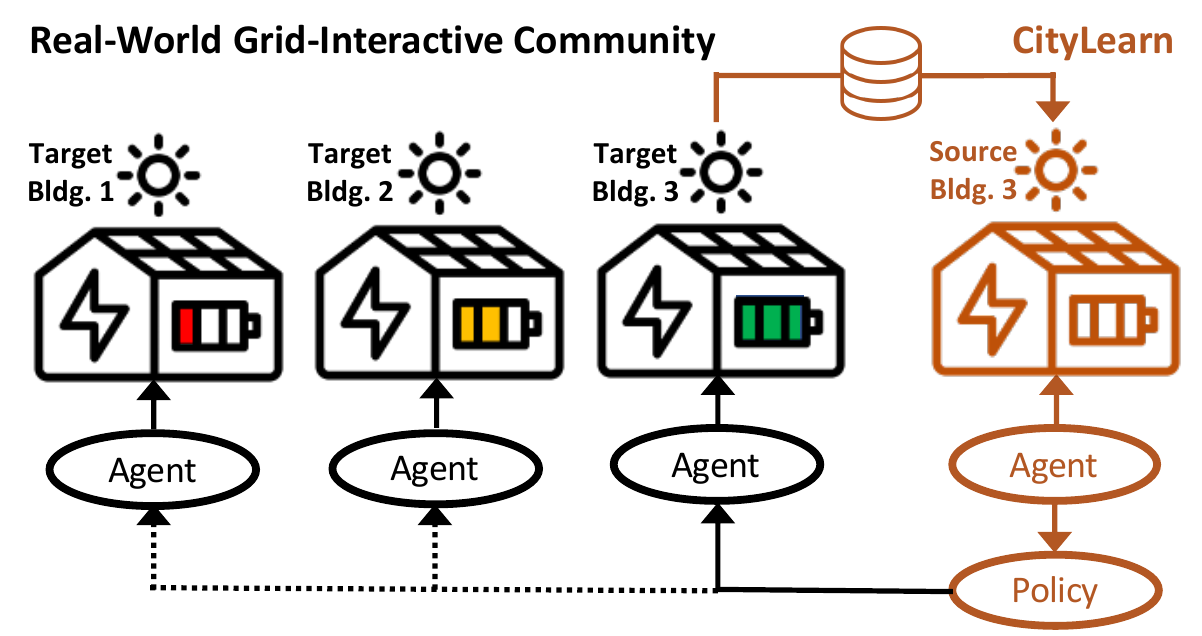}
        \caption{Deployment Strategy 2 \& 3 -- limited number of buildings and either at least one year (DS.2) or five months (DS.3) of data are used to create limited number of policies, hence the need for transfer learning when deployed in the grid-interactive community. For a community of 3 buildings, the trained policy for building 3 (source) is transferred to building N (targets) where N$\le$3.}
         \label{fig:deployment_strategy_2_and_3}
    \end{subfigure}
    \caption{Deployment strategies differing in spatial and temporal data availability.}
    \label{fig:deployment_strategy}
\end{figure}

Depicted in \cref{fig:deployment_strategy_1}, in deployment strategy 1 (DS.1), at least one year of simulated or measured data from each building is available and used for agent training in the digital twin. Building energy system specifications are also known. The aggregator develops a CityLearn environment model of the community with the provided data and trains agents that control each building's battery independently on the one-year dataset. DS.1 depicts a best-case scenario whereby the aggregator faces no temporal limitations in the available data as one year of smart meter captures seasonal variations in occupant behaviour and weather conditions. Likewise, there is no spatial limitation as all buildings' batteries will be controlled by agents that were trained on building-specific samples. In our implementation, one year of data for each building is used in training each building's policy for 10 episodes. The trained policies are then deployed in their respective real-world buildings and simulated using a deterministic policy for one episode.

\subsubsection{Deployment Strategy 2 (DS.2)} \label{sec:implementation-deployment_strategies-strategy_2}
Depicted in \cref{fig:deployment_strategy_2_and_3}, DS.2 is similar to DS.1 in the sense that the temporal dimension of training data that is known captures seasonal variations. However, only a limited number of customers provide their data. Still, all customers will like to benefit from an optimized home battery control system. This data limitation issue is a common situation where building data may not be readily available or privacy concerns hinder data sharing, and alludes to a transfer learning problem in which the policies that are learned from limited building data (source buildings) will be deployed and evaluated in an uncharted environment (target buildings). In our implementation, we run an experiment for each building as the source building and other 16 buildings, as well as the source building itself, as target buildings. After training the source building agent's policy on one year of data and for 10 episodes, the policy is transferred to the target buildings and updated with new samples in the target buildings for one year and one episode.

\subsubsection{Deployment Strategy 3 (DS.3)} \label{sec:implementation-deployment_strategies-strategy_3}
DS.3 builds upon DS.2; however, the temporal data coverage is limited and agents must train on partial knowledge of the buildings' seasonal loads. We limit the data availability to the first five months (August 1-December 31, 2016) of the measured time series data. As done in DS.2, we run an experiment for each building as the source building and other 16 buildings, as well as the source building itself, as target buildings. After training the source building agent policy for 10 episodes, it is transferred to the target buildings and updated with new samples in the target buildings for the remaining seven months (January 1-July 31, 2017) in one episode.


\subsection{Key Performance Indicators} \label{sec:implementation-kpi}
We evaluate the agents' performance on seven KPIs that are to be minimized: electricity consumption, $D$; electricity price, $C$; carbon emissions, $G$; zero net energy, $Z$; average daily peak, $P$; ramping, $R$ and (1 - Load Factor), ($1 - L$) for a total of $n$ time steps in an episode. $P$, $R$ and ($1 - L$) are grid-level KPIs that are calculated using the aggregated district-level hourly net electricity consumption (kWh), $E_h^{\textrm{district}}$. $E$, $C$, $G$ and $Z$ are building-level KPIs that are calculated using the building-level hourly net electricity consumption (kWh), $E_h^{\textrm{building}}$, and are reported at the grid level as the average of the building-level values.

Electricity consumption, $D$, is defined in \cref{eqn:kpi-electricity_consumption} as the sum of only non-negative $E_h^{\textrm{building}}$ as the objective is to minimize the energy consumed but not profit from the excess generation.

\begin{equation}
    D = \sum_{h=0}^{n-1}{\textrm{max} \left (0,E_h^{\textrm{building}} \right)}
    \label{eqn:kpi-electricity_consumption}
\end{equation}

Electricity price, $C$, is defined in \cref{eqn:kpi-electricity_price} as the sum of non-negative building-level net electricity price, $E_h^{\textrm{building}} \times T_h$ (\$), where $T_h$ is the electricity rate at hour $h$ that is specified in \cref{tab:elecricity_rate}.

\begin{equation}
    C = \sum_{h=0}^{n-1}{\textrm{max} \left (0,E_h^{\textrm{building}} \times T_h \right )}
    \label{eqn:kpi-electricity_price}
\end{equation}

Carbon emissions, $G$, is defined in \cref{eqn:kpi-carbon_emissions} as the sum of building-level carbon emissions (kg\textsubscript{CO\textsubscript{2}e}), $E_h^{\textrm{building}} \times O_h$, where $O_h$ is the carbon intensity (kg\textsubscript{CO\textsubscript{2}e}/kWh) at hour $h$.

\begin{equation}
    G = \sum_{h=0}^{n-1}{\textrm{max} \left (0,E_h^{\textrm{building}} \times O_h \right )}
    \label{eqn:kpi-carbon_emissions}
\end{equation}

Zero net energy, $Z$, is defined in \cref{eqn:kpi-zero_net_energy} as the sum of both negative and positive values of $E_h^{\textrm{building}}$.

\begin{equation}
    Z = \sum_{h=0}^{n-1}E_h^{\textrm{building}}
    \label{eqn:kpi-zero_net_energy}
\end{equation}

Average daily peak, $P$, is defined in \cref{eqn:kpi-average_daily_peak} as the mean of the daily maximum $E_h^{\textrm{district}}$ where $d$ is the day of year index.

\begin{equation}
    P = \frac{
        \mathlarger{\sum}_{d=0}^{364} \mathlarger{\sum}_{h=0}^{23} {\textrm{max} \left (E_{24d + h}^{\textrm{district}}, \dots, E_{24d + 23}^{\textrm{district}} \right)}
    }{365}
    \label{eqn:kpi-average_daily_peak}
\end{equation}

Ramping, $R$ is defined in \cref{eqn:kpi-ramping} as the absolute difference of consecutive $E_h^{\textrm{district}}$. It represents the smoothness of the district’s load profile where low $R$ means there is gradual increase in grid load even after self-generation becomes unavailable in the evening and early morning. High $R$ means abrupt change in grid load that may lead to unscheduled strain on grid infrastructure and blackouts as a result of supply deficit.

\begin{equation}
    R = \sum_{h=0}^{n-1}  \lvert E_{h}^{\textrm{district}} - E_{h - 1}^{\textrm{district}} \rvert
    \label{eqn:kpi-ramping}
\end{equation}

Load factor, $L$ is defined in \cref{eqn:kpi-load_factor} as the average ratio of monthly average and peak $E_{h}^{\textrm{district}}$ where $m$ is the month index. $L$ is the efficiency of electricity consumption and is bounded between 0 (very inefficient) and 1 (highly efficient) thus, the goal is to maximize $L$ or minimize $1 - L$.

\begin{equation}
    1 - L = \mathlarger{\mathlarger{ \Bigg ( }}
        \mathlarger{\mathlarger{\sum}}_{m=0}^{11} 1 - \frac{
            \left (
                \sum_{h=0}^{729} E_{730m + h}^{\textrm{district}}
            \right ) \div 730
        }{
            \textrm{max} \left (E_{730m}^{\textrm{district}}, \dots, E_{730m + 729}^{\textrm{district}} \right )
    } \mathlarger{\mathlarger{ \Bigg ) }} \div 12
    \label{eqn:kpi-load_factor}
\end{equation}

For the remainder of the paper, the KPIs are reported as normalized values with respect to the baseline outcome (\cref{eqn:kpi-normalization}) where the baseline outcome is when buildings are not equipped with batteries i.e., no control.

\begin{equation}
    \textrm{KPI} = \frac{{\textrm{KPI}\textsubscript{control}}}{\textrm{KPI}\textsubscript{baseline (no battery)}}
    \label{eqn:kpi-normalization}
\end{equation}

\section{Results} \label{sec:results}
\subsection{Exploratory Data Analysis}
\begin{figure*}
    \centering
    \begin{subfigure}[]{0.675\textwidth}
        \centering
        \includegraphics[width=\textwidth]{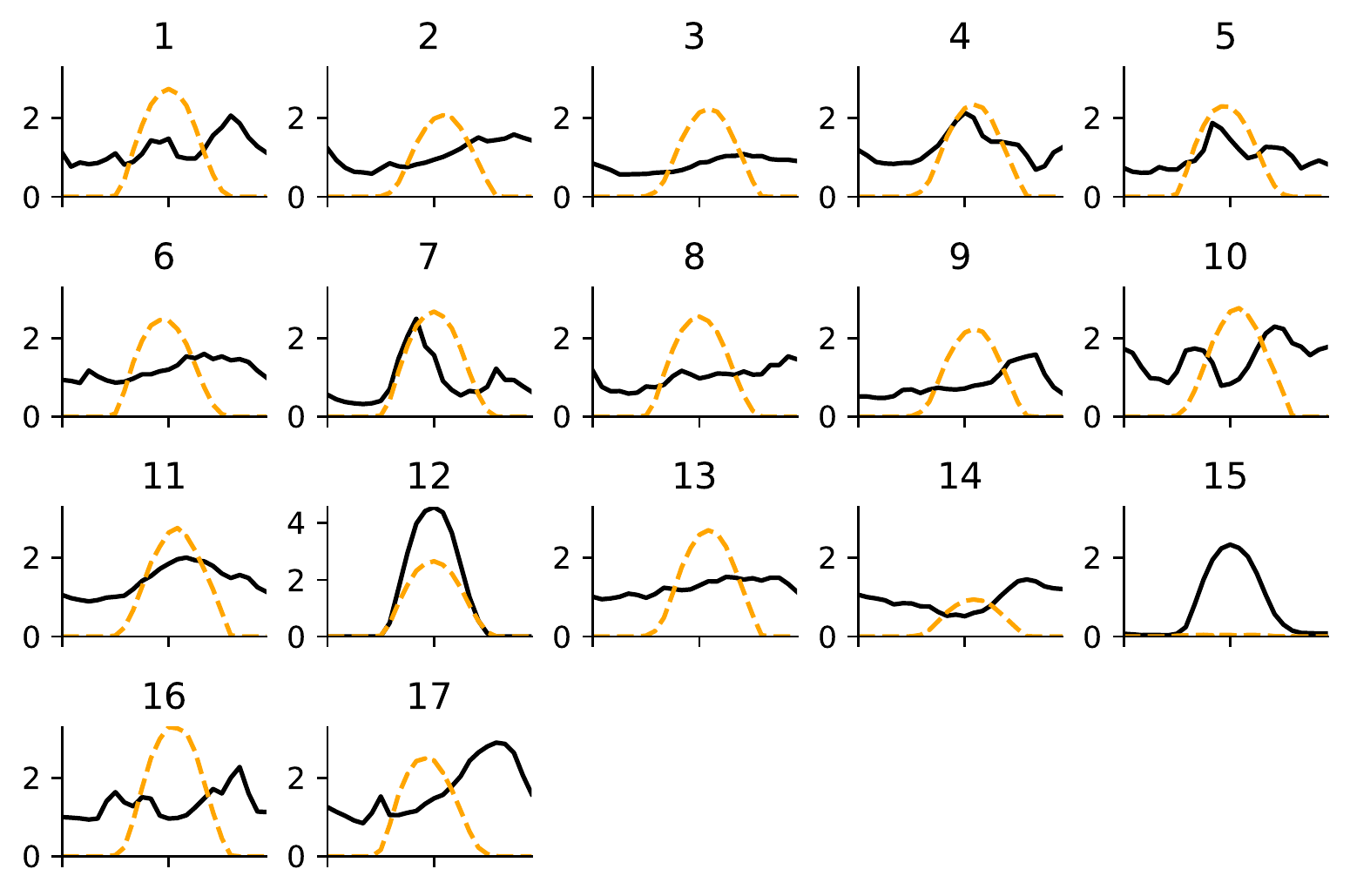}
        \caption{Building-level.}
        \label{fig:building_average_daily_load_and_generation_profile}
    \end{subfigure}
    \begin{subfigure}[]{0.3\textwidth}
        \centering
        \includegraphics[width=\textwidth]{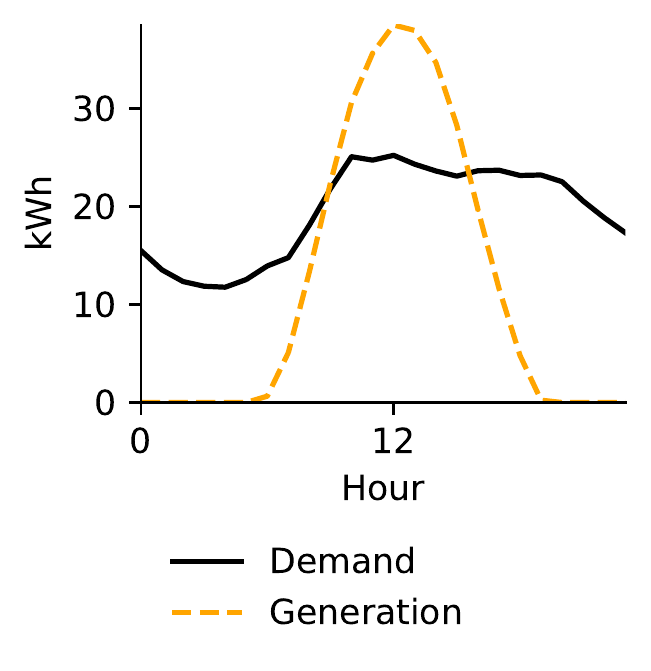}
        \caption{District-level.}
        \label{fig:district_average_daily_load_and_generation_profile}
    \end{subfigure}
    \caption{Average daily electricity demand without battery for flexibility (solid black line) and solar generation (dashed orange line) profiles. Note that building 12 has a different y-scale from the other buildings.}
    \label{fig:average_daily_load_and_generation_profile}
\end{figure*}

\cref{fig:average_daily_load_and_generation_profile} shows the average daily electricity demand (black line) and solar generation (dotted orange line) profiles at the building-level (\cref{fig:building_average_daily_load_and_generation_profile}) and aggregated sum district-level (\cref{fig:district_average_daily_load_and_generation_profile}). It shows the variability in demand across buildings as each building exhibits a unique shape and profile. The buildings also experience peak load at different times of the day on average with some buildings peaking at midday and others later in the evening. These observations of the building demand indicate diversity in occupant behaviour patterns and highlight the need for adaptive control strategies that are able to learn the unique building energy demand patterns. 

The time of peak generation does not coincide with the time of peak load in over half of the buildings. Building 15, although installed with PV system, does not generate electricity. Thus, the RL control agents must learn to charge the batteries during periods of surplus self-generation and discharge during peak hours but also learn to select adequate actions in the absence of supplemental renewable on-site generation.

The district-level profiles shows early district peak before noon and surplus generation that can harnessed for load shifting later in the evening. Also, there is a rapid ramping occurring before the peak that can be reduced by discharging stored energy at that time.

\subsection{Deployment Strategy 1}
\subsubsection{Load Profiles}
\begin{figure*}
    \centering
    \begin{subfigure}[]{0.675\textwidth}
        \centering
        \includegraphics[width=\textwidth]{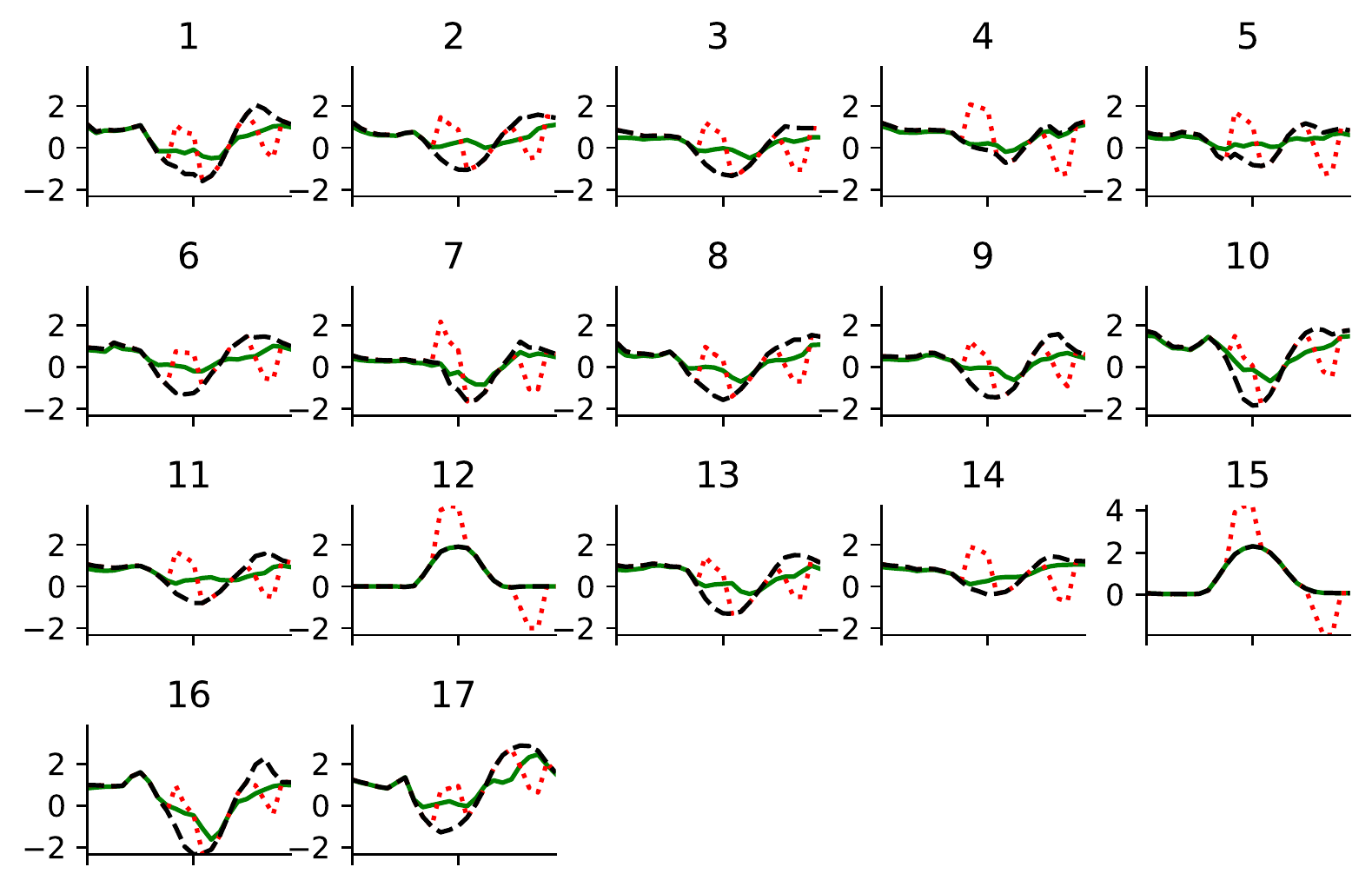}
        \caption{Building-level.}
        \label{fig:deployment_strategy_1_0_building_average_daily_load_profile}
    \end{subfigure}
    \begin{subfigure}[]{0.3\textwidth}
        \centering
        \includegraphics[width=\textwidth]{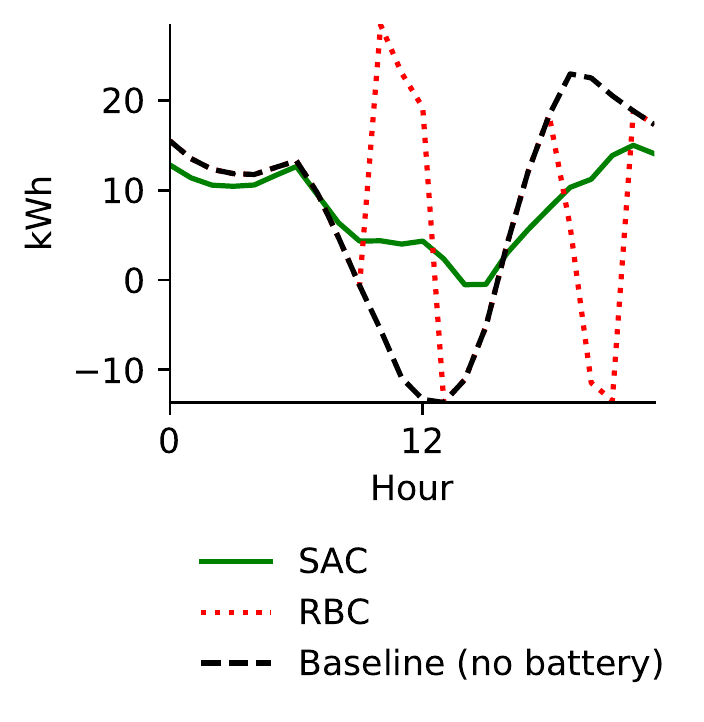}
        \caption{District-level.}
        \label{fig:deployment_strategy_1_0_district_average_daily_load_profile}
    \end{subfigure}
    \caption{Average daily net-electricity consumption for baseline (dashed black line), reference RBC (dotted red line) and SAC (solid green line) controls for last episode of DS.1. Note that building 15 has a different y-scale from the other buildings.}
    \label{fig:deployment_strategy_1_0_average_daily_load_profile}
\end{figure*}

\cref{fig:deployment_strategy_1_0_average_daily_load_profile} shows the resultant average daily net-electricity consumption profiles for the reference RBC and SAC control policies compared to the baseline (no battery) profile in the final deterministic control episode of DS.1 at the building-level (\cref{fig:deployment_strategy_1_0_building_average_daily_load_profile}) and district-level (\cref{fig:deployment_strategy_1_0_district_average_daily_load_profile}. The RBC policy causes unintended global peaks in most buildings, as well as in the district, and local peaks in other buildings before midday during the period of charge actions. These peaks deviate from the typical early to late evening peaks of residential-use buildings and can lead to grid instability without adequate planning. The RBC control policy also abruptly discharges stored energy later in the evening, which causes an abrupt ramp-up once the battery has been completely depleted. At the district level, the RBC control policy results in a higher peak compared to the baseline thus, results in worse-off district level energy flexibility. The SAC control policy improves the load shape in almost all buildings and the aggregated district where the SAC profiles have lower peaks compared to the RBC and baseline, are smoother with no rapid ramping and take advantage of excess solar generation in the afternoon. The exceptions to this improvement are seen in buildings 4, 7, 12 and 15 where the SAC policy has an approximately the same profile as the baseline thus, indicates that the use of the advanced controller for electrical storage has no effect on the buildings' energy flexibility. Buildings 4 and 7 originally have early peak before generation is added (see \cref{fig:building_average_daily_load_and_generation_profile}) and lower demand after noon. Hence, there is not a lot of load to offset from discharging the battery to cause significant change in the consumption profile.

\begin{figure}
    \begin{subfigure}[]{\columnwidth}
        \centering
        \includegraphics[width=\columnwidth]{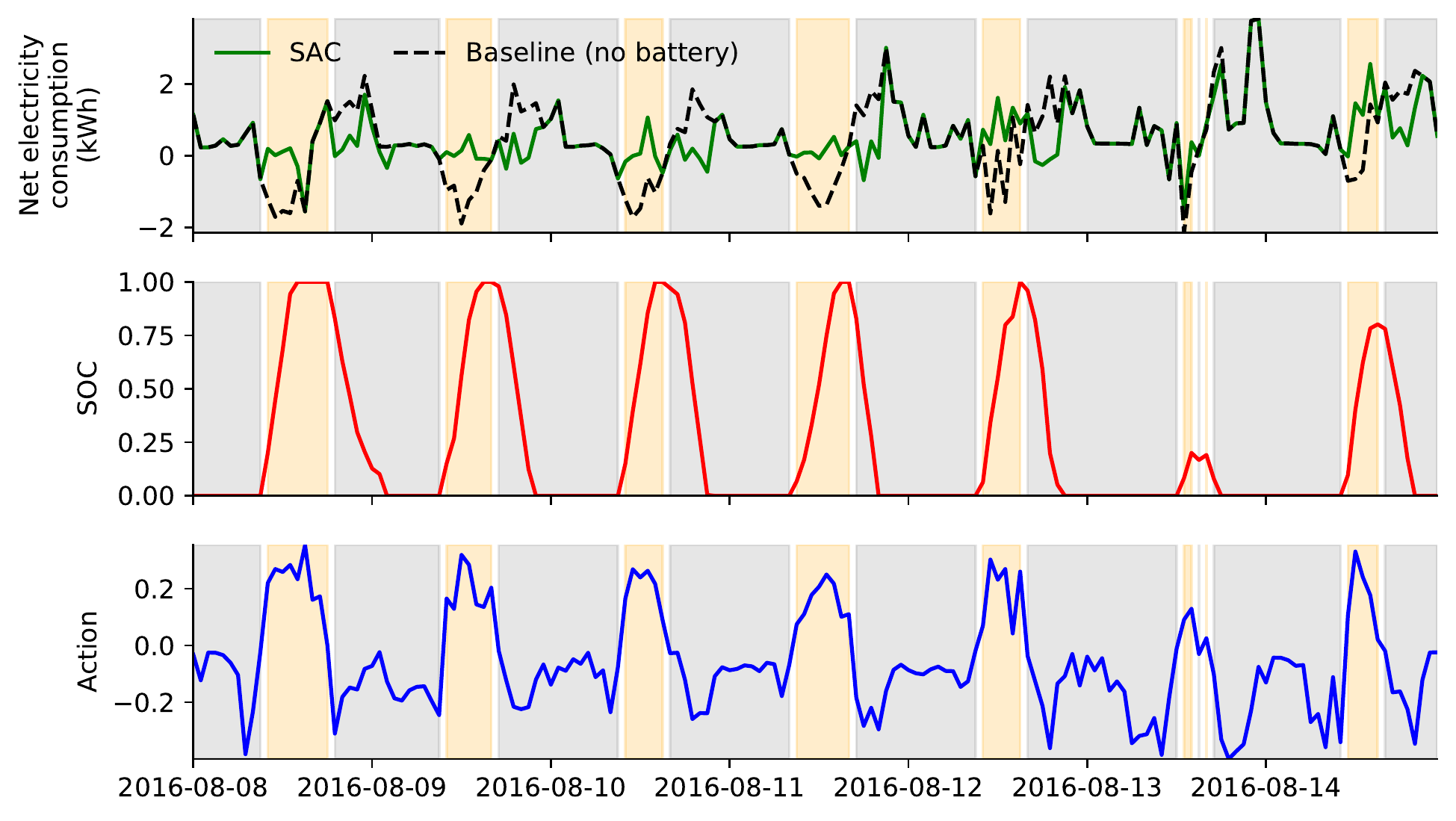}
        \caption{Building 2.}
        \label{fig:deployment_strategy_1_0_building_2_profile_snapshot}
    \end{subfigure}
    \begin{subfigure}[]{\columnwidth}
        \centering
        \includegraphics[width=\columnwidth]{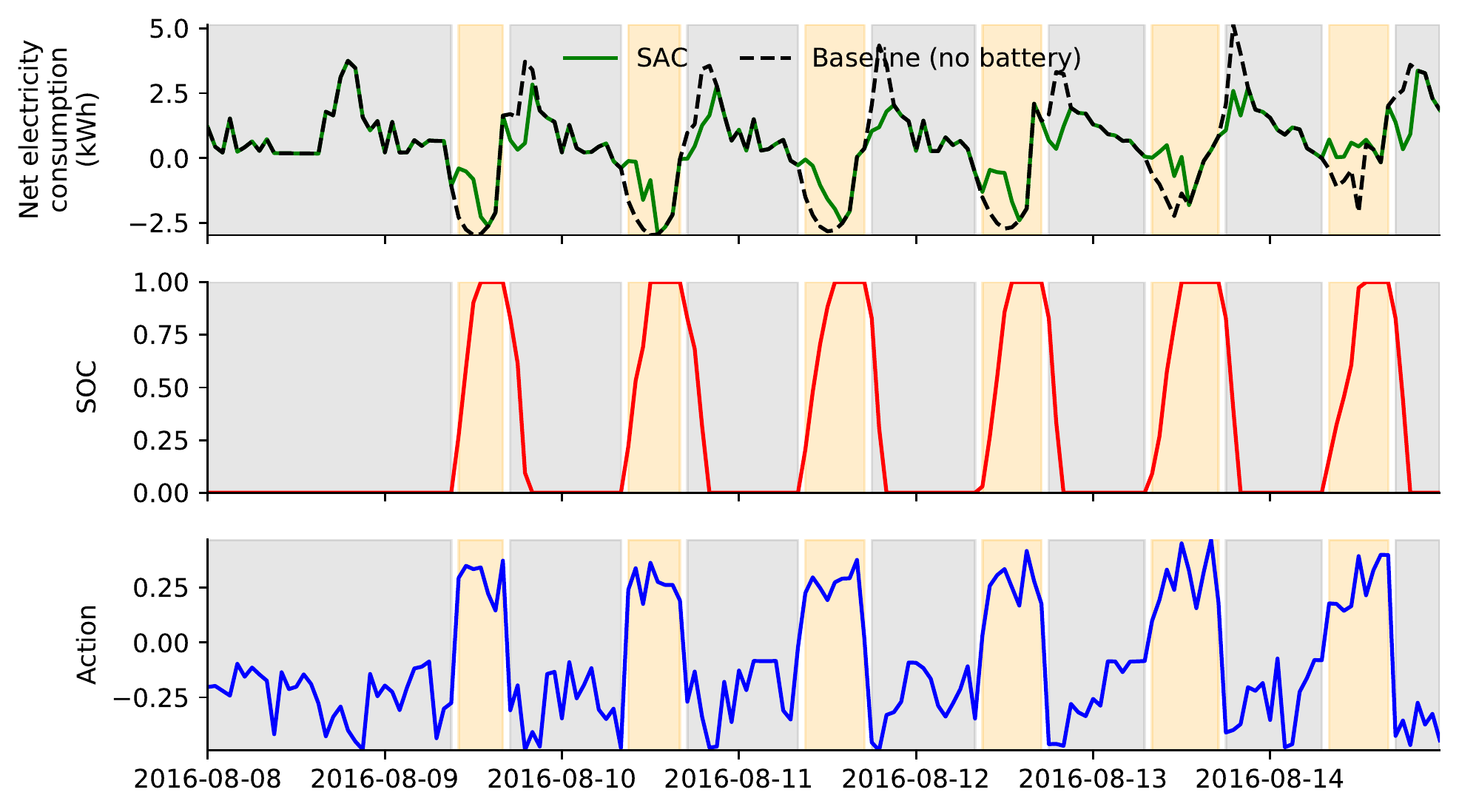}
        \caption{Building 7.}
        \label{fig:deployment_strategy_1_0_building_7_profile_snapshot}
    \end{subfigure}
    \begin{subfigure}[]{\columnwidth}
        \centering
        \includegraphics[width=\columnwidth]{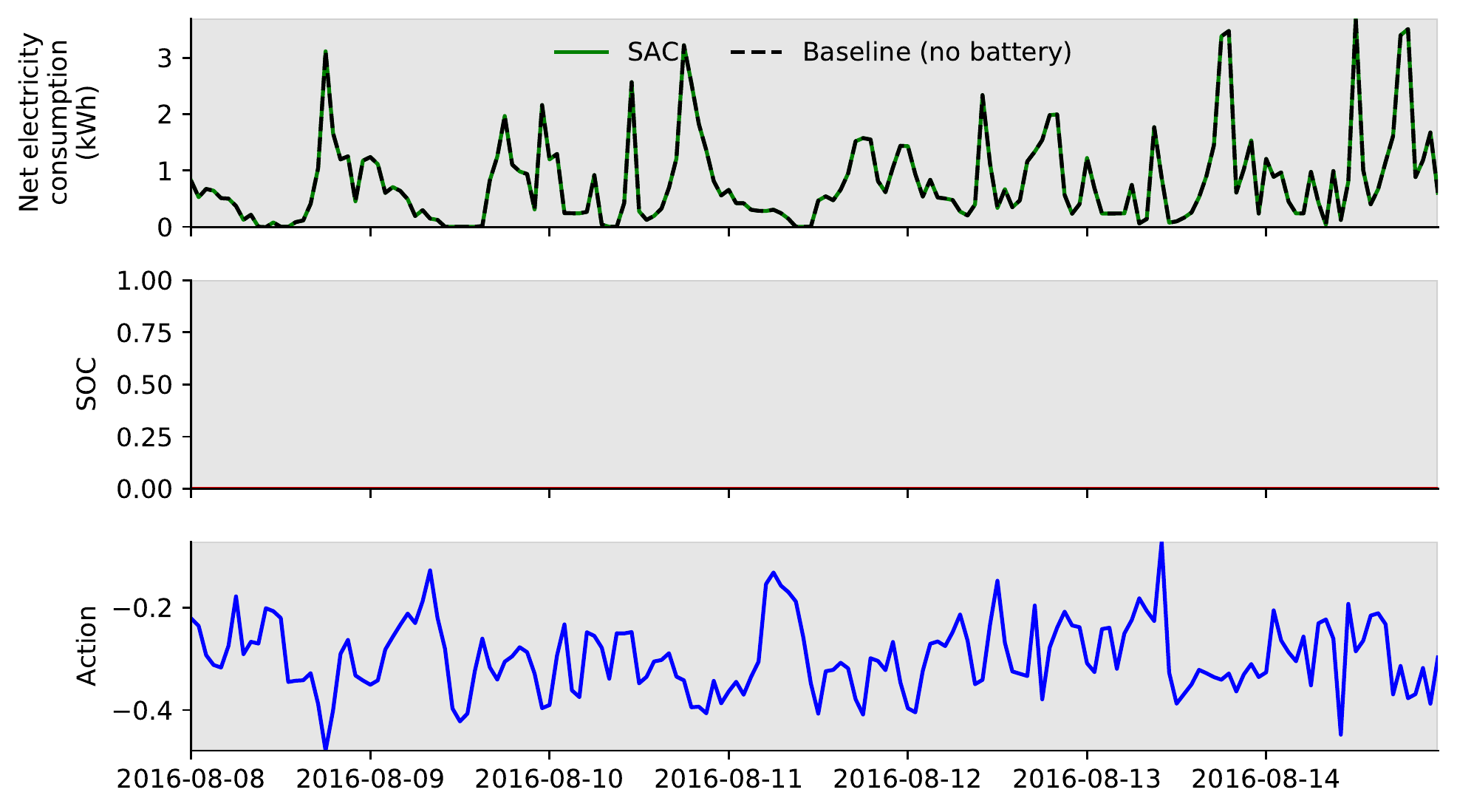}
        \caption{Building 15.}
        \label{fig:deployment_strategy_1_0_building_15_profile_snapshot}
    \end{subfigure}
    \caption{Net-electricity consumption for SAC (solid green line) and baseline (dashed black line) control, battery SOC (solid red line) and SAC agent action (solid blue line) for a week in the final deterministic control episode of DS.1. The grey and orange background fills indicate discharge and charge actions respectively.}
    \label{fig:deployment_strategy_1_0_building_profile_snapshot}
\end{figure}

\cref{fig:deployment_strategy_1_0_building_profile_snapshot} shows a snapshot of full week's net-electricity consumption for SAC control policy (solid green line) compared to the baseline (dashed black line), battery SOC (solid red line) and SAC agent action (solid blue line)  in the final deterministic control episode of DS.1 for buildings 2 (\cref{fig:deployment_strategy_1_0_building_2_profile_snapshot}), 7 (\cref{fig:deployment_strategy_1_0_building_7_profile_snapshot}) and 15 (\cref{fig:deployment_strategy_1_0_building_15_profile_snapshot}) as these buildings are representative of the diversity of profiles seen in all buildings. The grey and orange background fills indicate discharge and charge actions respectively. In buildings 2 and 7, the agent calls for charging within the late morning and late afternoon when electricity rate is at the lowest (see \cref{tab:elecricity_rate}) and discharges later in the day. Although the agents fail to learn when the maximum SOC has been reached and still take charging actions at SOC=1, it is a conservative policy as immediate discharging actions could be detrimental to later periods of high peaks and expensive electricity rate. The independent SAC agents also learn the unique building energy needs indicated by wider time ranges of discharging in building 2 compared to 7 in the last 2 days. In contrast to buildings 2 and 7, building 15, maintains a completely drained battery and calls for discharge action throughout the shown period. The implication is that the SAC and baseline control strategies yield the same net-electricity consumption profiles. Recall from \cref{fig:building_average_daily_load_and_generation_profile} that building 15 does not generate electricity hence, will only use electricity from the grid to charge its battery. The observed agent actions in building 15 is conservative as it avoids expensive electricity consumption from the grid in the absence of self-generation that will be needed to charge its battery.

\subsubsection{Key Performance Indicators}
\begin{figure*}
    \centering
    \begin{subfigure}[]{0.55\textwidth}
        \centering
        \includegraphics[width=\textwidth]{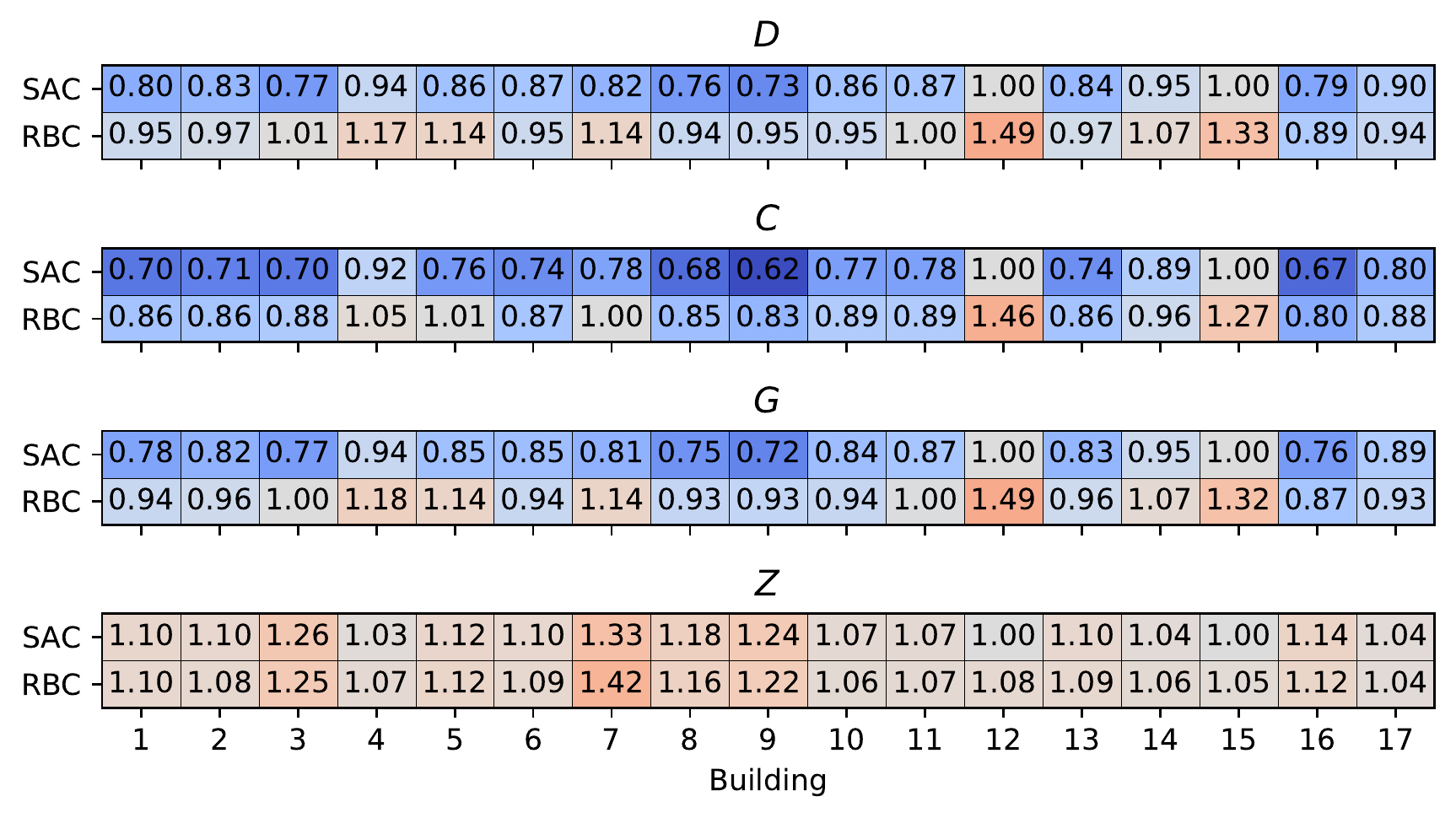}
        \caption{Building-level KPIs.}
        \label{fig:deployment_strategy_1_0_building_cost_summary}
    \end{subfigure}
    \begin{subfigure}[]{0.425\textwidth}
        \centering
        \includegraphics[width=\textwidth]{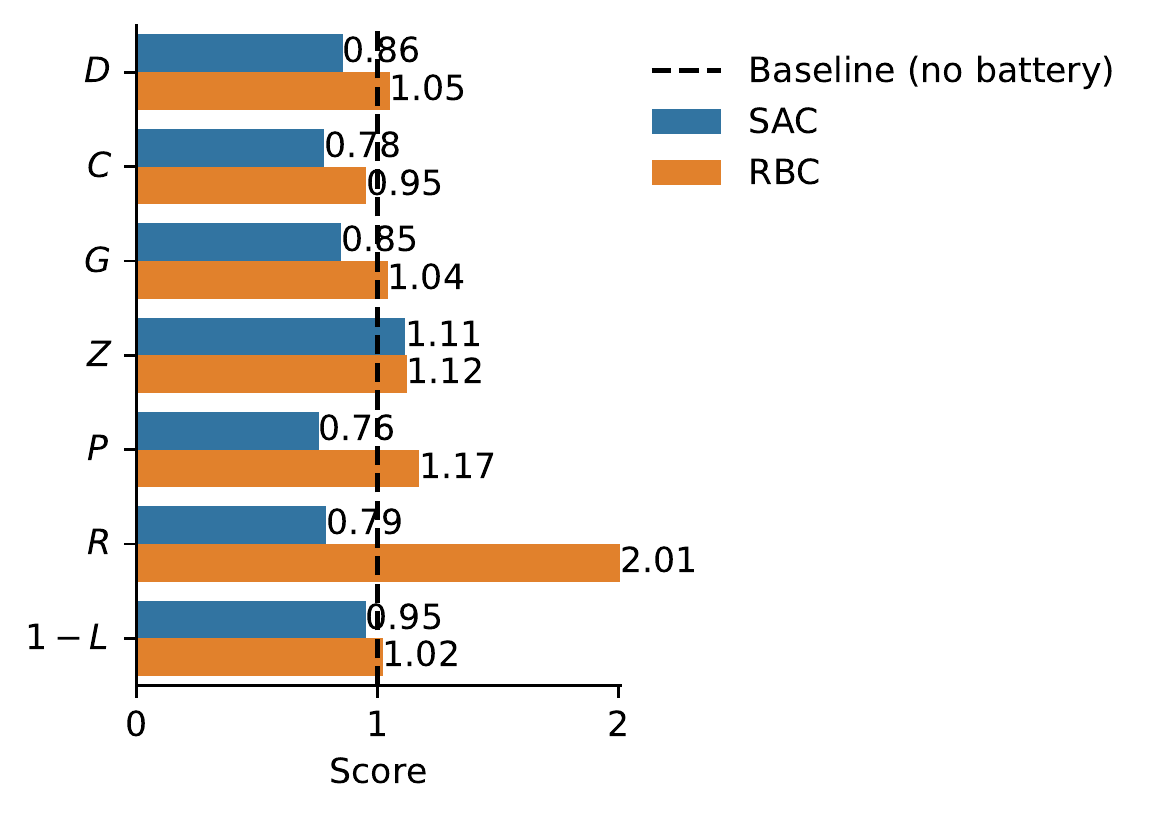}
        \caption{District-level KPIs.}
        \label{fig:deployment_strategy_1_0_district_cost_summary}
    \end{subfigure}
    \caption{Electricity consumption ($D$), electricity price ($C$), carbon emissions ($G$), zero net energy ($Z$), average daily peak ($P$), ramping ($R$) and 1 - load factor ($1 - L$) KPIs for the baseline, RBC and SAC control policies in the final deterministic control episode of DS.1.}
    \label{deployment_strategy_1_0_cost_summary}
\end{figure*}

The KPIs for the baseline, RBC and SAC control policies in the final deterministic control episode of DS.1 are shown in \cref{deployment_strategy_1_0_cost_summary}. For the building-level KPIs (\cref{fig:deployment_strategy_1_0_building_cost_summary}), red and blue cells indicate better and worse performance compared to the baseline. The SAC policy minimizes the electricity consumption ($D$), electricity price ($C$) and carbon emissions ($G$) KPIs compared to the baseline and RBC policy. In contrast, both the SAC and RBC policies worsen the zero net energy ($Z$) KPI compared to the baseline. While the RBC policy also yields lower $Z$ compared to the SAC policy in 9/17 buildings, the advantage the RBC provides is minuscule in the order of $\leq 0.1$. In buildings 12 and 15, $D$, $C$ and $G$ are worse-off under the control of the RBC policy  and unchanged under the control of the SAC architecture compared to the baseline.

The district-level KPIs (\cref{fig:deployment_strategy_1_0_district_cost_summary}) show that on average, the building-level scores with the exception of $Z$ are improved by over 15\% when the batteries in the district use an advanced independent SAC policy compared to a simplified expert RBC policy system, and are better than the baseline KPIs. Despite the zero net energy ($Z$) KPI being worse than the baseline on average, other district-level KPIs (average daily peak ($P$), ramping ($R$) and 1 - load factor ($1 - L$)) are minimized when compared to the baseline and RBC even without any coordination nor cooperation between buildings. Quantitatively, $R$, $P$ and $1 -L$ are reduced by over 50\%, 25\% and 5\% when the all buildings in the district use independent SAC policies in place of RBCs to manage electricity storage.

\subsection{Deployment Strategy 2}
\subsubsection{Key Performance Indicators after Transfer Learning}

\begin{figure*}
    \begin{subfigure}[]{\textwidth}
        \centering
        \includegraphics[width=\textwidth]{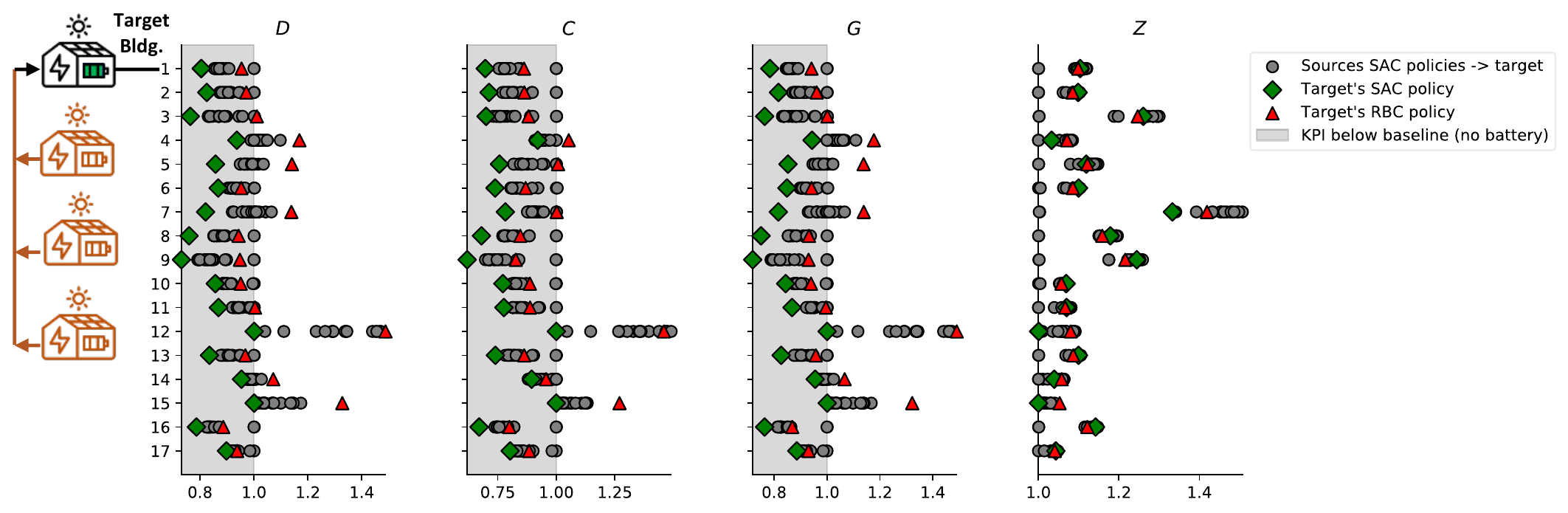}
        \caption{Building-level KPIs when one of other 16 buildings' (sources) agents is transferred to a target building.}
        \label{fig:deployment_strategy_2_0_target_building_cost_summary}
    \end{subfigure}
    \begin{subfigure}[]{\textwidth}
        \centering
        \includegraphics[width=\textwidth]{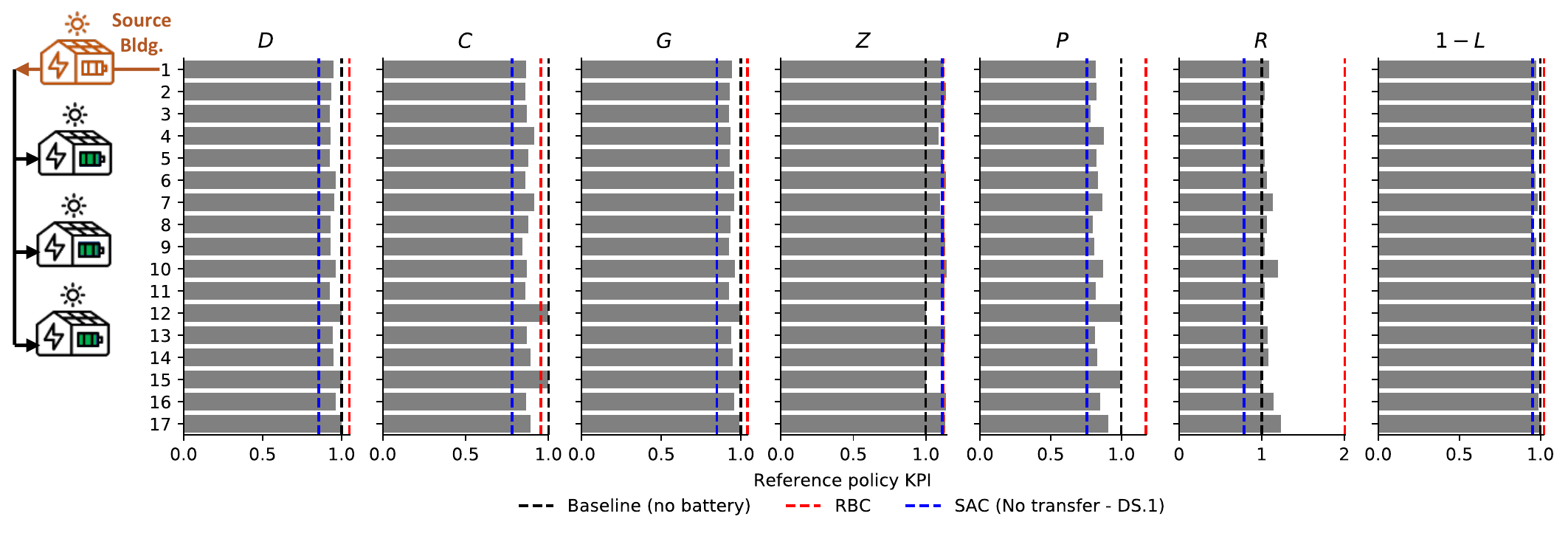}
        \caption{District-level KPIs when a source building's SAC agent is transferred to other 16 buildings (targets).}
        \label{fig:deployment_strategy_2_0_district_cost_summary}
    \end{subfigure}
    \caption{Electricity consumption ($D$), electricity price ($C$), carbon emissions ($G$), zero net energy ($Z$), average daily peak ($P$), ramping ($R$) and 1 - load factor ($1 - L$) KPIs after transfer learning in DS.2.}
    \label{fig:deployment_strategy_2_0_cost_summary}
\end{figure*}

In \cref{fig:deployment_strategy_2_0_cost_summary}, we summarize the KPIs after transfer learning has occurred in DS.2 at the building-level (\cref{fig:deployment_strategy_2_0_target_building_cost_summary}) and district-level (\cref{fig:deployment_strategy_2_0_district_cost_summary}). At the building-level, we make comparisons amongst the KPIs achieved when a target building is controlled by its own SAC policy, its own RBC policy or the SAC policy of one of the other buildings when the other building is used as a source building. Electricity consumption ($D$), electricity price ($C$) and carbon emissions ($G$) KPIs show similar distributions compared to zero net energy ($Z$) KPI. Target buildings experience worse performance in all four building-level KPIs when inheriting policies from source buildings compared to the performance of the target's own SAC policy. However, the performance remain better than that of the target's own RBC and the baseline in most cases of $D$, $C$ and $G$ KPIs. $Z$ is at or above the baseline irrespective of transfer learning. In summary, we find that compared to the KPIs as a result of a target building's own SAC policy, $D$, $C$ and $G$ increase by 10\%-12\% on average in all buildings whereas, $Z$ increases by 3.5\% on average in 5/17 target buildings and decreases by 1.7\% on average in the remaining 12/17 target buildings, when a source building's policy is deployed in a target building.

For a given district-level KPI, the values are comparable when a source building's SAC agent is transferred to other buildings (targets) irrespective of what building is used as the source as seen in \cref{fig:deployment_strategy_2_0_district_cost_summary}. Generally, KPIs become worse in the transfer learning scenario compared to the no-transfer learning scenario (DS.1) but in most cases, are lower than those achieved when an RBC or no control policy is used. Compared to the KPIs when each building uses it's own trained agent (no transfer -- DS.1), Quantitatively, the average change in $D$, $C$, $G$, $Z$, $P$, $R$ and $1 - L$ when compared to the no-transfer scenario is 9.6\%$\pm$2.6\%, 11.0\%$\pm$4.5\%, 10.2\%$\pm$2.5\%, -0.2\%$\pm$4.4\%, 9.8\%$\pm$6.2\%, 28.7\%$\pm$6.8\% and 2.6\%$\pm$1.6\% respectively where $Z$ and $1-L$ are least affected on average.

\subsection{Deployment Strategy 3}
\subsubsection{Key Performance Indicators without transfer learning}
\begin{figure*}
    \centering
    \begin{subfigure}[]{0.55\textwidth}
        \centering
        \includegraphics[width=\textwidth]{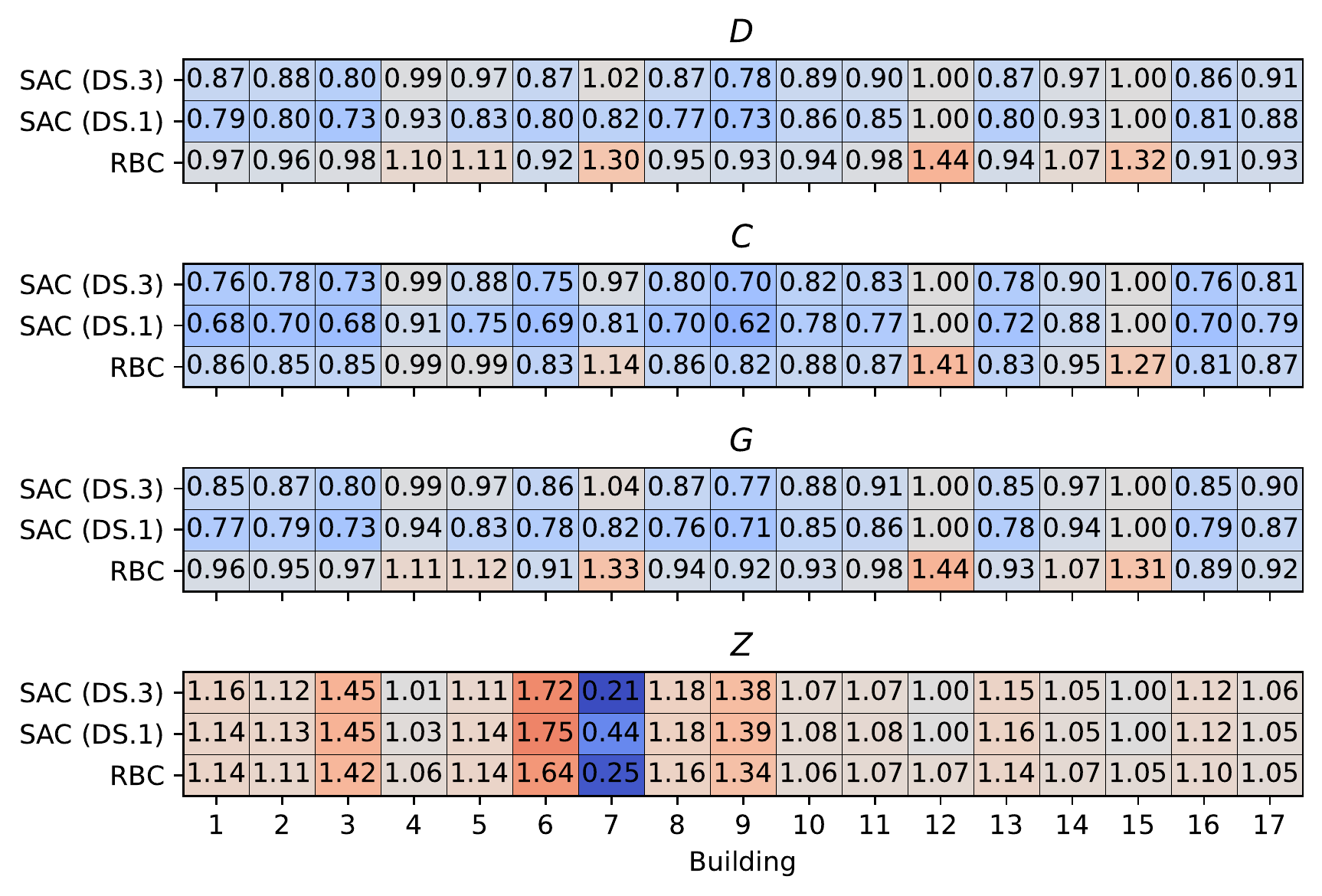}
        \caption{Building-level KPIs.}
        \label{fig:deployment_strategy_3_1_building_cost_summary}
    \end{subfigure}
    \begin{subfigure}[]{0.425\textwidth}
        \centering
        \includegraphics[width=\textwidth]{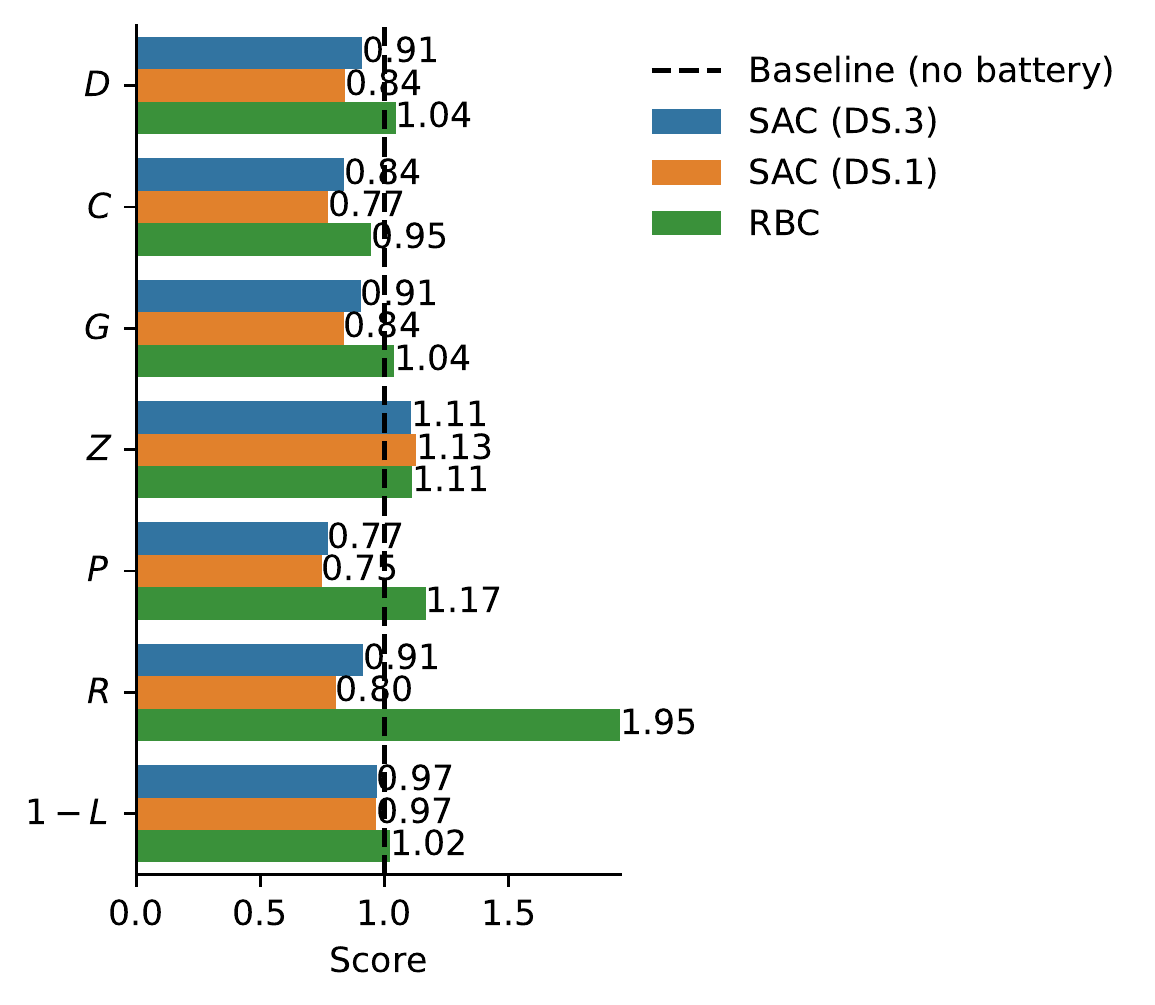}
        \caption{District-level KPIs.}
        \label{fig:deployment_strategy_3_1_district_cost_summary}
    \end{subfigure}
    \caption{Electricity consumption ($D$), electricity price ($C$), carbon emissions ($G$), zero net energy ($Z$), average daily peak ($P$), ramping ($R$) and 1 - load factor ($1 - L$) KPIs for unseen seven-month period in DS.3. The KPIs are evaluated for the baseline scenario (no battery), RBC policy, SAC policies that have been trained on all 12 months of data (DS.1) and SAC policies that have been trained on just the initial 5 months of data (DS.3).}
    \label{deployment_strategy_3_1_cost_summary}
\end{figure*}

\cref{deployment_strategy_3_1_cost_summary} shows building-level (\cref{fig:deployment_strategy_3_1_building_cost_summary}) and district-level (\cref{fig:deployment_strategy_3_1_district_cost_summary}) KPIs for the unseen seven-month period in DS.3. The KPIs are evaluated for the baseline scenario (no battery), RBC policy, SAC policies that have been trained on all 12 months of data (DS.1) and SAC policies that have been trained on just the initial 5 months of data (DS.3). Although, training the SAC agents on shorter time period of data slightly worsens the electricity consumption ($D$), electricity price ($C$) and carbon emissions ($G$) by 6.7\%, 6.4\% and 6.9\% respectively, the KPIs are still generally better than those of the baseline and reference RBC as seen in (\cref{fig:deployment_strategy_3_1_district_cost_summary}). In contrast, zero net energy ($Z$) KPI is improved by 2.0\% on average where in 6 buildings the values are approximately the same irrespective of training data length. Ramping ($R$) on the district-level is worsened by 11.0
\% when the SAC agents are trained on just 5 months of data while average daily peak ($P$) worsens by just 2.0\%, and 1 - load factor ($1 - L$) remains unchanged as shown in \cref{fig:deployment_strategy_3_1_district_cost_summary}. These results suggest that the full seasonality in occupant behaviour and weather need not be available in the training data to achieve comparable performance.


\subsubsection{Key Performance Indicators after Transfer Learning}
\begin{figure*}
    \begin{subfigure}[]{\textwidth}
        \centering
        \includegraphics[width=\textwidth]{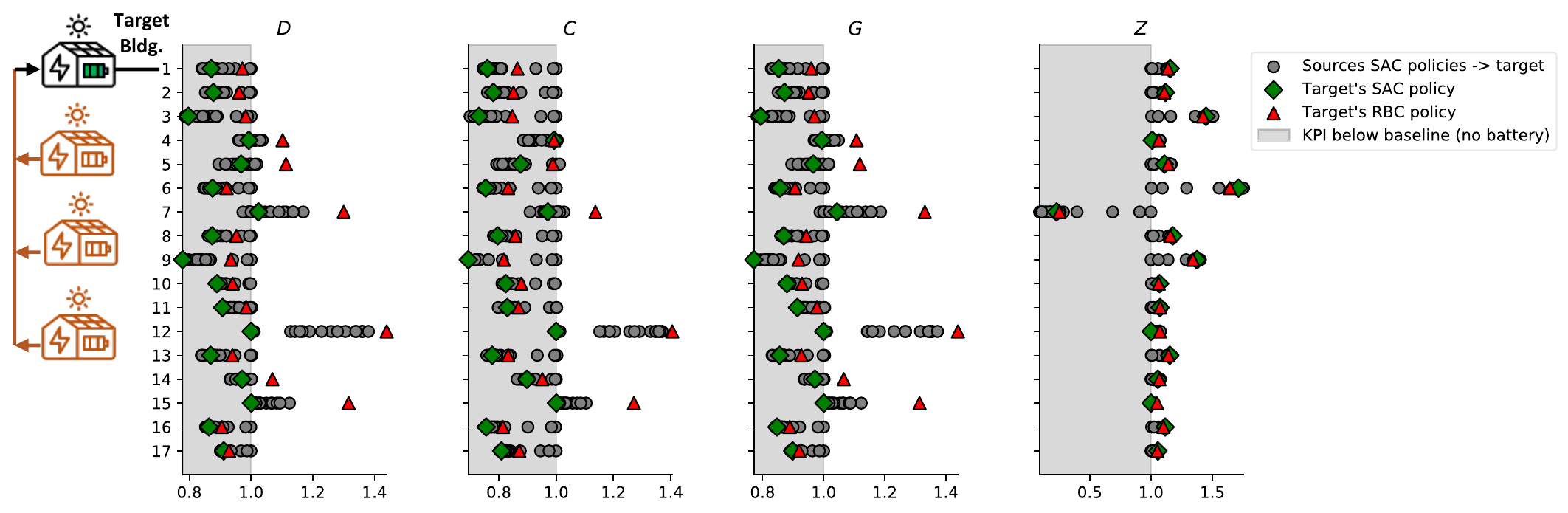}
        \caption{Building-level KPIs when one of other 16 buildings' (sources) agents is transferred to a target building.}
        \label{fig:deployment_strategy_3_2_target_building_cost_summary}
    \end{subfigure}
    \begin{subfigure}[]{\textwidth}
        \centering
        \includegraphics[width=\textwidth]{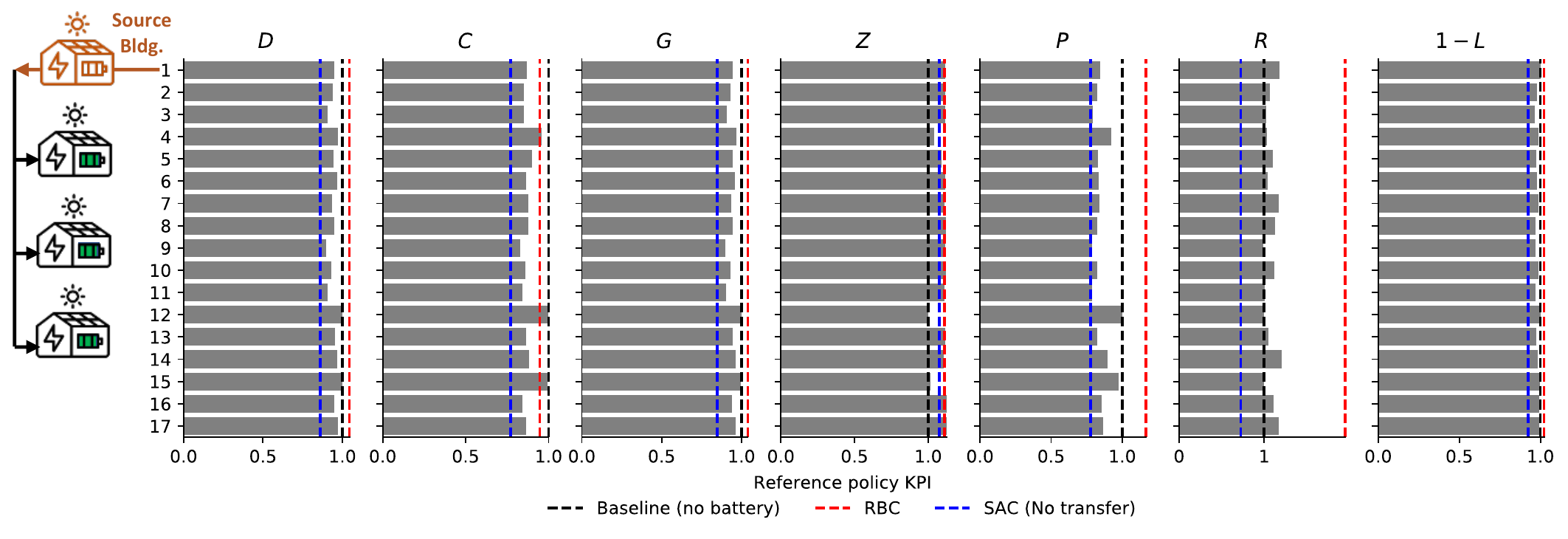}
        \caption{District-level KPIs when a source building's SAC agent is transferred to other 16 buildings (targets).}
        \label{fig:deployment_strategy_3_2_district_cost_summary}
    \end{subfigure}
    \caption{Electricity consumption ($D$), electricity price ($C$), carbon emissions ($G$), zero net energy ($Z$), average daily peak ($P$), ramping ($R$) and 1 - load factor ($1 - L$) KPIs after transfer learning in DS.3.}
    \label{fig:deployment_strategy_3_2_cost_summary}
\end{figure*}

Similar to \cref{fig:deployment_strategy_2_0_cost_summary} for DS.2, in \cref{fig:deployment_strategy_3_2_cost_summary}, we summarize the KPIs after transfer learning in DS.3 at the building-level (\cref{fig:deployment_strategy_3_2_target_building_cost_summary}) and district-level (\cref{fig:deployment_strategy_3_2_district_cost_summary}). $D$, $C$ and $G$ for DS.3 are similar to those achieved in DS.2 except when the target building is building 7 where in DS.3, more source building to target building transfers have KPIs that are worse-off compared to the baseline. Meanwhile, $Z$ is improved in the case of a number of source building policies being transferred to building 7 as shown in \cref{fig:deployment_strategy_3_2_target_building_cost_summary}. In summary, we find that compared to the KPIs as a result of a target building's own SAC policy, $D$, $C$, $G$ and $Z$ increase by 4.0\%-6.0\% ($\approx$ half the change for DS.2) or decrease by 0.1\%-3.8\% on average when a source building's policy is deployed in a target building.

The district-level KPIs for DS.3 after transfer learning are similar to those observed in DS.2 where generally, KPIs are worsened if the rest of the district inherits the policy of a source building. Quantitatively, the average change in $D$, $C$, $G$, $Z$, $P$, $R$ and $1 - L$ when compared to the no-transfer scenario is 8.8\%$\pm$2.9\%, 11.3\%$\pm$5.1\%, 9.8\%$\pm$2.8\%, 2.1\%$\pm$3.9\%, 7.7\%$\pm$6.2\%, 35.5\%$\pm$7.1\% and 5.8\%$\pm$1.1\% respectively where the changes are similar to those found in DS.2. Also, $Z$ and $1-L$ are least affected on average as seen in DS.2.



\section{Discussion} \label{sec:discussion}
\subsection{Real-world Challenges}
The three barriers hindering the adoption of RL in grid-interactive buildings discussed by \cite{wang2020reinforcement} and addressed in this work are corroborated by the real-world challenges for RL in grid-interactive introduced by \citeauthor{NWEYE2022100202} in \cite{NWEYE2022100202}. Challenge 1 -- \textit{Being able to learn on live systems from limited samples}, is analogous to the barrier of time-consuming and data-intensive training requirements in \cite{wang2020reinforcement} as it relates to the use of fewer data to achieve reasonable comparable performance to a case where there is ample training data. The proposed DS.3 strategy that shows comparable performance in using five months of data for training instead of a full year's 12 months of data that captures all seasonal variations addresses this challenge.

Challenge 4 -- \textit{Reasoning about system constraints that should never or rarely be violated} is analogous to the control security and robustness problem highlighted in \cite{wang2020reinforcement} that is concerned with maintaining comfortable occupant conditions while under the influence of the RL controller as well as evaluating the RL controller on its ability to provide resiliency-based actions such as ensuring that the storage systems SOCs are never completely depleted. Challenge 8 -- \textit{Training off-line from the fixed logs of an external behavior policy} touches on the training data scarcity problem where the controller may learn from samples of different lengths generated by a reference expert system controller e.g. RBC but also touches on the control security problem where the control policy of a expert system that is industry-accepted that does not violate system constraints can be used to initialize the training of the RL controller. The modification of the SAC agent used in this work, where instead of random actions, an RBC policy provides actions during exploration in offline training seeks to provides a viable and practical solution to Challenges 4 and 8. Also, the backup controller in CityLearn ensures that comfort is always satisfied while providing flexibility. 

 Challenge 5 -- \textit{Interacting with systems that are partially observable, which can alternatively be viewed as systems that are non-stationary or stochastic} alludes to the generalizability problem discussed in \cite{wang2020reinforcement} as the controller should be able to provide good performance in the event of perturbations to the environment upon which it has been train on or in the event that it is transferred to a new environment with different a different domain or target space. The transfer learning results in DS.2 and DS.3 in this work show that independently trained SAC agents can provide good performance when transferred to an environments they were not trained on. Thus, as new developments occur in the neighborhood, or as more homes are fitted with HEMS, the lack of historical data can be alleviated by using existing building control model to jump-start the use of the available DERs efficiently.

\subsection{Occupant Centric-Control}
The OCC framework suggest to integrate occupant preferences and behavior into the controller design \cite{park2019critical}. In our work, we use the building loads as proxy for occupant satisfaction and assume that maintaining the building loads as measured while providing flexibility, would result in the highest possible occupant comfort, because there is no negative impact on the occupant. Here, we demonstrate that focusing on individual buildings can lead to positive affects such as peak shifting, while advanced controllers can provide flexibility and grid support on the district level without negatively impacting the individual buildings. As such, our work unifies the research areas of occupant-centric-control and grid-interactive buildings, by offering a new perspective to harness energy flexibility on the district level by capitalizing on the uniqueness of buildings due to diverse occupant behavior. This is achieved using the unique CityLearn environment which can be easily extended with novel data-driven approaches for occupant behavior\cite{Quintana2022}. It is important to note that this outcome depends on the size of the battery and the PV systems, and the climate where the neighborhood is located as these factors can influence how much flexibility can be harnessed. Thus, future work should investigate the effect of different storage and PV capacities, as well as the influence of climatic conditions, on energy flexibility.

While the DERs used in this work are limited to only batteries for active storage and PV for self-generation, other resources are possible e.g., thermal storage systems to offset space cooling, space heating and domestic hot water loads, thermal mass for pre-cooling and pre-heating and EVs. Also, the fixed demand approach used in our MERLIN approach can be adjusted to allow for load shedding by means of thermostatically controlled loads e.g. heat-pump for space cooling and heating. To this end, \citeauthor{pinto2021data} proposed a framework that allows for partial cooling load satisfaction by a heat pump that satisfies comfort constraints through the use of LSTM architecture to capture the internal building temperature dynamics in the CityLearn environment.

\subsection{Cooperative Multi-Agent Control}
We use an independent control architecture in our case-study implementation of MERLIN where each building's battery is controlled by a isolated control policy that does not take other buildings actions and states into account. This approach preserves occupancy privacy and control security but may limit the full grid-level flexibility potential as uncooperative control can cause adverse effects such as peak shifting instead of reduction. \citeauthor{Vazquez-Canteli2020} developed the MARLISA agent, a cooperative multi-agent reinforcement learning agent that inherits the properties of the SAC algorithm used in our work but also includes an internal model that predicts the district net electricity consumption for use as an environment state \cite{Vazquez-Canteli2020}. Their results showed that increased flexibility compared to independent agents can be achieved when agents are aware of the impact of actions taken by other agents in a district. Likewise, \citeauthor{pinto2022enhancing} demonstrated that better performance can be achieved by sharing district level information and designing their objective function to promote cooperation \cite{pinto2022enhancing}. Thus, a future research implementation of MERLIN is to introduce the idea of cooperation within the grid-interactive community.

\subsection{Reference RBC Design}
The choice of expert RBC policy in use determines the extent to which an advanced control system such as RL provides in building control applications \cite{NWEYE2022100202}. In this work, we utilize the RBC policy that was implemented in the real-world policy as a reference for an RL policy and show superior performance of the RL approach. However, the RBC policy in use may not be the most ideal for the unique occupant behaviours in the community. Thus, the need to use RBC polices that are optimized for the specific systems they control as a starting reference. Such level of specificity in RBC definition can be expensive to arrive at and will need to be re-evaluated periodically to account for perturbations in the controlled system.

\subsection{Policy Explainability}
One of the challenges of real-world applications of RL in GEBs is the explainability of policies (Challenge 9 in \cite{dulac2021challenges}). Simplification of complex RL algorithms through rule extraction for use as sequence-of-operation or framing as a supervised learning problem e.g. decision tree \citeauthor{Xilei2022}, can help accelerate its adoption in real-world building operations as it reduces adoption risk, increases reliability and communicates clearly, the controller actions by providing a familiar control \textit{language} that is understood by most building stakeholders (e.g. homeowners, tenants, building managers, policy makers, e.t.c) while achieving performance that is comparable to the black-box RL alternative. Therefore, future work should investigate the feasibility of rule extraction in the current implementation of MERLIN for DER control.
\section{Conclusion} \label{sec:conclusion}
Our work addresses the data requirement, control security and generalizability challenges hindering real-world adoption of RL in real-world building control applications through the proposed MERLIN framework. With a 17-ZNE building dataset from a real-world grid-interactive community in the CityLearn environment, we demonstrate that:

\begin{enumerate}
    \item Independent RL-controllers for batteries improve building and district level KPIs compared to the baseline ZNE buildings and reference RBC by tailoring their policies to individual building characteristics (DS.1).
    \item Despite unique occupant behaviours, transferring the RL policy of any one of the buildings to other buildings provides comparable performance to a policy trained on building-specific samples, while reducing the cost of training (DS.2 and DS.3).
    \item Training the RL-controllers on limited temporal data that does not capture full seasonality in occupant behaviour has little effect on performance when the controller is deployed in the same building that it was trained on or transferred to other buildings (DS.3).
    \item Although, the ZNE condition of the buildings could be maintained or worsened as a result of controlled batteries, KPIs that are typically improved by ZNE condition (electricity price and carbon emissions) are further improved when the batteries are managed by advanced controllers that learn to charge the batteries at times of low monetary and environmental cost (DS.1, DS.2 and DS.3).
\end{enumerate}

Overall, we unify the areas of occupant-centric and grid-interactive buildings and lay the foundation for further studies. Future work should investigate the inclusion of thermostatically controlled DERs, effect of differently sized DERs across buildings, the impact of climate on generalizability and policy explainability in the current implementation of MERLIN for DER control.


\bibliographystyle{model1-num-names}

\bibliography{references}

\begin{thebibliography}{59}
\expandafter\ifx\csname natexlab\endcsname\relax\def\natexlab#1{#1}\fi
\providecommand{\url}[1]{\texttt{#1}}
\providecommand{\href}[2]{#2}
\providecommand{\path}[1]{#1}
\providecommand{\DOIprefix}{doi:}
\providecommand{\ArXivprefix}{arXiv:}
\providecommand{\URLprefix}{URL: }
\providecommand{\Pubmedprefix}{pmid:}
\providecommand{\doi}[1]{\href{http://dx.doi.org/#1}{\path{#1}}}
\providecommand{\Pubmed}[1]{\href{pmid:#1}{\path{#1}}}
\providecommand{\bibinfo}[2]{#2}
\ifx\xfnm\relax \def\xfnm[#1]{\unskip,\space#1}\fi
\bibitem[{Energy Information Administration(2022)}]{eia-mer-00352211}
Energy Information Administration, \bibinfo{title}{November 2022 monthly energy
  review}, \bibinfo{year}{2022}. \URLprefix
  \url{https://www.eia.gov/totalenergy/data/monthly/archive/00352211.pdf}.
\bibitem[{Goldstein et~al.(2020)Goldstein, Gounaridis, and
  Newell}]{goldstein2020carbon}
\bibinfo{author}{B.~Goldstein}, \bibinfo{author}{D.~Gounaridis},
  \bibinfo{author}{J.~P. Newell},
\newblock \bibinfo{title}{The carbon footprint of household energy use in the
  united states},
\newblock \bibinfo{journal}{Proceedings of the National Academy of Sciences}
  \bibinfo{volume}{117} (\bibinfo{year}{2020}) \bibinfo{pages}{19122--19130}.
\bibitem[{Leibowicz et~al.(2018)Leibowicz, Lanham, Brozynski, Vazquez-Canteli,
  Castejon, and Nagy}]{Lanham2018}
\bibinfo{author}{B.~D. Leibowicz}, \bibinfo{author}{C.~M. Lanham},
  \bibinfo{author}{M.~T. Brozynski}, \bibinfo{author}{J.~R. Vazquez-Canteli},
  \bibinfo{author}{N.~C. Castejon}, \bibinfo{author}{Z.~Nagy},
\newblock \bibinfo{title}{{Optimal decarbonization pathways for urban
  residential building energy services}},
\newblock \bibinfo{journal}{Applied Energy} \bibinfo{volume}{230}
  (\bibinfo{year}{2018}) \bibinfo{pages}{1311--1325}.
\bibitem[{{Yekini Suberu} et~al.(2014){Yekini Suberu}, {Wazir Mustafa}, and
  Bashir}]{YEKINISUBERU2014499}
\bibinfo{author}{M.~{Yekini Suberu}}, \bibinfo{author}{M.~{Wazir Mustafa}},
  \bibinfo{author}{N.~Bashir},
\newblock \bibinfo{title}{Energy storage systems for renewable energy power
  sector integration and mitigation of intermittency},
\newblock \bibinfo{journal}{Renewable and Sustainable Energy Reviews}
  \bibinfo{volume}{35} (\bibinfo{year}{2014}) \bibinfo{pages}{499--514}.
\bibitem[{Gowrisankaran et~al.(2016)Gowrisankaran, Reynolds, and
  Samano}]{doi:10.1086/686733}
\bibinfo{author}{G.~Gowrisankaran}, \bibinfo{author}{S.~S. Reynolds},
  \bibinfo{author}{M.~Samano},
\newblock \bibinfo{title}{Intermittency and the value of renewable energy},
\newblock \bibinfo{journal}{Journal of Political Economy} \bibinfo{volume}{124}
  (\bibinfo{year}{2016}) \bibinfo{pages}{1187--1234}.
\bibitem[{Neukomm et~al.(2019)Neukomm, Nubbe, and Fares}]{osti_1508212}
\bibinfo{author}{M.~Neukomm}, \bibinfo{author}{V.~Nubbe},
  \bibinfo{author}{R.~Fares},
\newblock \bibinfo{title}{Grid-interactive efficient buildings}
  (\bibinfo{year}{2019}).
\bibitem[{Nikzad and Mozafari(2014)}]{NIKZAD201483}
\bibinfo{author}{M.~Nikzad}, \bibinfo{author}{B.~Mozafari},
\newblock \bibinfo{title}{Reliability assessment of incentive- and priced-based
  demand response programs in restructured power systems},
\newblock \bibinfo{journal}{International Journal of Electrical Power \& Energy
  Systems} \bibinfo{volume}{56} (\bibinfo{year}{2014}) \bibinfo{pages}{83--96}.
\bibitem[{Haider et~al.(2016)Haider, See, and Elmenreich}]{HAIDER2016166}
\bibinfo{author}{H.~T. Haider}, \bibinfo{author}{O.~H. See},
  \bibinfo{author}{W.~Elmenreich},
\newblock \bibinfo{title}{A review of residential demand response of smart
  grid},
\newblock \bibinfo{journal}{Renewable and Sustainable Energy Reviews}
  \bibinfo{volume}{59} (\bibinfo{year}{2016}) \bibinfo{pages}{166--178}.
\bibitem[{Gelazanskas and Gamage(2014)}]{Gelazanskas2014DemandDirection}
\bibinfo{author}{L.~Gelazanskas}, \bibinfo{author}{K.~A.~A. Gamage},
\newblock \bibinfo{title}{{Demand side management in smart grid: A review and
  proposals for future direction}},
\newblock \bibinfo{journal}{Sustainable Cities and Society}
  \bibinfo{volume}{11} (\bibinfo{year}{2014}) \bibinfo{pages}{22--30}.
\bibitem[{Burger et~al.(2017)Burger, Chaves-{\'A}vila, Batlle, and
  P{\'e}rez-Arriaga}]{BURGER2017395}
\bibinfo{author}{S.~Burger}, \bibinfo{author}{J.~P. Chaves-{\'A}vila},
  \bibinfo{author}{C.~Batlle}, \bibinfo{author}{I.~J. P{\'e}rez-Arriaga},
\newblock \bibinfo{title}{A review of the value of aggregators in electricity
  systems},
\newblock \bibinfo{journal}{Renewable and Sustainable Energy Reviews}
  \bibinfo{volume}{77} (\bibinfo{year}{2017}) \bibinfo{pages}{395--405}.
\bibitem[{Drgoňa et~al.(2020)Drgoňa, Arroyo, Cupeiro~Figueroa, Blum, Arendt,
  Kim, Oll{\'{e}}, Oravec, Wetter, Vrabie, and Helsen}]{Drgona2020}
\bibinfo{author}{J.~Drgoňa}, \bibinfo{author}{J.~Arroyo},
  \bibinfo{author}{I.~Cupeiro~Figueroa}, \bibinfo{author}{D.~Blum},
  \bibinfo{author}{K.~Arendt}, \bibinfo{author}{D.~Kim}, \bibinfo{author}{E.~P.
  Oll{\'{e}}}, \bibinfo{author}{J.~Oravec}, \bibinfo{author}{M.~Wetter},
  \bibinfo{author}{D.~L. Vrabie}, \bibinfo{author}{L.~Helsen},
\newblock \bibinfo{title}{{All you need to know about model predictive control
  for buildings}},
\newblock \bibinfo{journal}{Annual Reviews in Control} \bibinfo{volume}{50}
  (\bibinfo{year}{2020}) \bibinfo{pages}{190--232}.
\bibitem[{Wang and Hong(2020)}]{Wang2020}
\bibinfo{author}{Z.~Wang}, \bibinfo{author}{T.~Hong},
\newblock \bibinfo{title}{{Reinforcement learning for building controls: The
  opportunities and challenges}},
\newblock \bibinfo{journal}{Applied Energy} \bibinfo{volume}{269}
  (\bibinfo{year}{2020}) \bibinfo{pages}{115036}.
\bibitem[{Privara et~al.(2011)Privara, {\v{S}}irok{\`y}, Ferkl, and
  Cigler}]{privara2011model}
\bibinfo{author}{S.~Privara}, \bibinfo{author}{J.~{\v{S}}irok{\`y}},
  \bibinfo{author}{L.~Ferkl}, \bibinfo{author}{J.~Cigler},
\newblock \bibinfo{title}{Model predictive control of a building heating
  system: The first experience},
\newblock \bibinfo{journal}{Energy and Buildings} \bibinfo{volume}{43}
  (\bibinfo{year}{2011}) \bibinfo{pages}{564--572}.
\bibitem[{Karlsson and Hagentoft(2011)}]{karlsson2011application}
\bibinfo{author}{H.~Karlsson}, \bibinfo{author}{C.-E. Hagentoft},
\newblock \bibinfo{title}{Application of model based predictive control for
  water-based floor heating in low energy residential buildings},
\newblock \bibinfo{journal}{Building and environment} \bibinfo{volume}{46}
  (\bibinfo{year}{2011}) \bibinfo{pages}{556--569}.
\bibitem[{Yuan and Perez(2006)}]{yuan2006multiple}
\bibinfo{author}{S.~Yuan}, \bibinfo{author}{R.~Perez},
\newblock \bibinfo{title}{Multiple-zone ventilation and temperature control of
  a single-duct vav system using model predictive strategy},
\newblock \bibinfo{journal}{Energy and buildings} \bibinfo{volume}{38}
  (\bibinfo{year}{2006}) \bibinfo{pages}{1248--1261}.
\bibitem[{Yang et~al.(2015)Yang, Nagy, Goffin, and Schlueter}]{Yang2015a}
\bibinfo{author}{L.~Yang}, \bibinfo{author}{Z.~Nagy},
  \bibinfo{author}{P.~Goffin}, \bibinfo{author}{A.~Schlueter},
\newblock \bibinfo{title}{{Reinforcement learning for optimal control of low
  exergy buildings}},
\newblock \bibinfo{journal}{Applied Energy} \bibinfo{volume}{156}
  (\bibinfo{year}{2015}) \bibinfo{pages}{577--586}.
\bibitem[{O'Neill et~al.(2010)O'Neill, Levorato, Goldsmith, and
  Mitra}]{o2010residential}
\bibinfo{author}{D.~O'Neill}, \bibinfo{author}{M.~Levorato},
  \bibinfo{author}{A.~Goldsmith}, \bibinfo{author}{U.~Mitra},
\newblock \bibinfo{title}{Residential demand response using reinforcement
  learning},
\newblock in: \bibinfo{booktitle}{2010 First IEEE international conference on
  smart grid communications}, \bibinfo{organization}{IEEE},
  \bibinfo{year}{2010}, pp. \bibinfo{pages}{409--414}.
\bibitem[{Liu and Henze(2006)}]{liu2006experimental}
\bibinfo{author}{S.~Liu}, \bibinfo{author}{G.~P. Henze},
\newblock \bibinfo{title}{Experimental analysis of simulated reinforcement
  learning control for active and passive building thermal storage inventory:
  Part 2: Results and analysis},
\newblock \bibinfo{journal}{Energy and buildings} \bibinfo{volume}{38}
  (\bibinfo{year}{2006}) \bibinfo{pages}{148--161}.
\bibitem[{Mozer(1998)}]{mozer1998neural}
\bibinfo{author}{M.~C. Mozer},
\newblock \bibinfo{title}{The neural network house: An environment hat adapts
  to its inhabitants},
\newblock in: \bibinfo{booktitle}{Proc. AAAI Spring Symp. Intelligent
  Environments}, volume~\bibinfo{volume}{58}, \bibinfo{year}{1998}.
\bibitem[{Costanzo et~al.(2016)Costanzo, Iacovella, Ruelens, Leurs, and
  Claessens}]{costanzo2016experimental}
\bibinfo{author}{G.~T. Costanzo}, \bibinfo{author}{S.~Iacovella},
  \bibinfo{author}{F.~Ruelens}, \bibinfo{author}{T.~Leurs},
  \bibinfo{author}{B.~J. Claessens},
\newblock \bibinfo{title}{Experimental analysis of data-driven control for a
  building heating system},
\newblock \bibinfo{journal}{Sustainable Energy, Grids and Networks}
  \bibinfo{volume}{6} (\bibinfo{year}{2016}) \bibinfo{pages}{81--90}.
\bibitem[{Ruelens et~al.(2016)Ruelens, Claessens, Quaiyum, De~Schutter,
  Babu{\v{s}}ka, and Belmans}]{ruelens2016reinforcement}
\bibinfo{author}{F.~Ruelens}, \bibinfo{author}{B.~J. Claessens},
  \bibinfo{author}{S.~Quaiyum}, \bibinfo{author}{B.~De~Schutter},
  \bibinfo{author}{R.~Babu{\v{s}}ka}, \bibinfo{author}{R.~Belmans},
\newblock \bibinfo{title}{Reinforcement learning applied to an electric water
  heater: From theory to practice},
\newblock \bibinfo{journal}{IEEE Transactions on Smart Grid}
  \bibinfo{volume}{9} (\bibinfo{year}{2016}) \bibinfo{pages}{3792--3800}.
\bibitem[{Drgoňa et~al.(2021)Drgoňa, Tuor, Chandan, and Vrabie}]{Drgona2021}
\bibinfo{author}{J.~Drgoňa}, \bibinfo{author}{A.~R. Tuor},
  \bibinfo{author}{V.~Chandan}, \bibinfo{author}{D.~L. Vrabie},
\newblock \bibinfo{title}{{Physics-constrained deep learning of multi-zone
  building thermal dynamics}},
\newblock \bibinfo{journal}{Energy and Buildings} \bibinfo{volume}{243}
  (\bibinfo{year}{2021}).
\bibitem[{Arroyo et~al.(2022)Arroyo, Manna, Spiessens, and Helsen}]{Arroyo2022}
\bibinfo{author}{J.~Arroyo}, \bibinfo{author}{C.~Manna},
  \bibinfo{author}{F.~Spiessens}, \bibinfo{author}{L.~Helsen},
\newblock \bibinfo{title}{{Reinforced model predictive control (RL-MPC) for
  building energy management}},
\newblock \bibinfo{journal}{Applied Energy} \bibinfo{volume}{309}
  (\bibinfo{year}{2022}) \bibinfo{pages}{118346}.
\bibitem[{Chen et~al.(2020)Chen, Cai, and Berg{\'{e}}s}]{Chen2020a}
\bibinfo{author}{B.~Chen}, \bibinfo{author}{Z.~Cai},
  \bibinfo{author}{M.~Berg{\'{e}}s},
\newblock \bibinfo{title}{{Gnu-RL: A Practical and Scalable Reinforcement
  Learning Solution for Building HVAC Control Using a Differentiable MPC
  Policy}},
\newblock \bibinfo{journal}{Frontiers in Built Environment} \bibinfo{volume}{6}
  (\bibinfo{year}{2020}) \bibinfo{pages}{316--325}.
\bibitem[{{Richard S. Sutton and Andrew G.
  Barto}(2018)}]{RichardS.SuttonandAndrewG.Barto2018a}
\bibinfo{author}{{Richard S. Sutton and Andrew G. Barto}},
  \bibinfo{title}{{Reinforcement Learning, Second Edition An Introduction}},
  \bibinfo{year}{2018}.
\bibitem[{Nagy et~al.(2018)Nagy, Park, and Vazquez-Canteli}]{Nagy2018}
\bibinfo{author}{Z.~Nagy}, \bibinfo{author}{J.~Y. Park},
  \bibinfo{author}{J.~Vazquez-Canteli},
\newblock \bibinfo{title}{{Reinforcement learning for intelligent environments:
  A Tutorial}},
\newblock in: \bibinfo{editor}{P.~Gardoni} (Ed.), \bibinfo{booktitle}{Handbook
  of Sustainable and Resilient Infrastructure}, \bibinfo{edition}{1} ed.,
  \bibinfo{publisher}{Routledge}, \bibinfo{year}{2018}.
\bibitem[{Mnih et~al.(2015)Mnih, Kavukcuoglu, Silver, Rusu, Veness, Bellemare,
  Graves, Riedmiller, Fidjeland, Ostrovski, Petersen, Beattie, Sadik,
  Antonoglou, King, Kumaran, Wierstra, Legg, and Hassabis}]{mnih15}
\bibinfo{author}{V.~Mnih}, \bibinfo{author}{K.~Kavukcuoglu},
  \bibinfo{author}{D.~Silver}, \bibinfo{author}{A.~A. Rusu},
  \bibinfo{author}{J.~Veness}, \bibinfo{author}{M.~G. Bellemare},
  \bibinfo{author}{A.~Graves}, \bibinfo{author}{M.~Riedmiller},
  \bibinfo{author}{A.~K. Fidjeland}, \bibinfo{author}{G.~Ostrovski},
  \bibinfo{author}{S.~Petersen}, \bibinfo{author}{C.~Beattie},
  \bibinfo{author}{A.~Sadik}, \bibinfo{author}{I.~Antonoglou},
  \bibinfo{author}{H.~King}, \bibinfo{author}{D.~Kumaran},
  \bibinfo{author}{D.~Wierstra}, \bibinfo{author}{S.~Legg},
  \bibinfo{author}{D.~Hassabis},
\newblock \bibinfo{title}{{Human-level control through deep reinforcement
  learning}},
\newblock \bibinfo{journal}{Nature} \bibinfo{volume}{518}
  (\bibinfo{year}{2015}) \bibinfo{pages}{529--533}.
\bibitem[{Silver et~al.(2017)Silver, Schrittwieser, Simonyan, Antonoglou,
  Huang, Guez, Hubert, Baker, Lai, Bolton, Chen, Lillicrap, Hui, and
  Sifre}]{Silver2017}
\bibinfo{author}{D.~Silver}, \bibinfo{author}{J.~Schrittwieser},
  \bibinfo{author}{K.~Simonyan}, \bibinfo{author}{I.~Antonoglou},
  \bibinfo{author}{A.~Huang}, \bibinfo{author}{A.~Guez},
  \bibinfo{author}{T.~Hubert}, \bibinfo{author}{L.~Baker},
  \bibinfo{author}{M.~Lai}, \bibinfo{author}{A.~Bolton},
  \bibinfo{author}{Y.~Chen}, \bibinfo{author}{T.~Lillicrap},
  \bibinfo{author}{F.~Hui}, \bibinfo{author}{L.~Sifre},
\newblock \bibinfo{title}{{Article Mastering the game of Go without human
  knowledge}},
\newblock \bibinfo{journal}{Nature} \bibinfo{volume}{550}
  (\bibinfo{year}{2017}) \bibinfo{pages}{354--359}.
\bibitem[{Vinyals et~al.(2019)Vinyals, Babuschkin, Czarnecki, Mathieu, Dudzik,
  Chung, Choi, Powell, Ewalds, Georgiev, Oh, Horgan, Kroiss, Danihelka, Huang,
  Sifre, Cai, Agapiou, Jaderberg, Vezhnevets, Leblond, Pohlen, Dalibard,
  Budden, Sulsky, Molloy, Paine, Gulcehre, Wang, Pfaff, Wu, Ring, Yogatama,
  W{\"{u}}nsch, McKinney, Smith, Schaul, Lillicrap, Kavukcuoglu, Hassabis,
  Apps, and Silver}]{Vinyals2019}
\bibinfo{author}{O.~Vinyals}, \bibinfo{author}{I.~Babuschkin},
  \bibinfo{author}{W.~M. Czarnecki}, \bibinfo{author}{M.~Mathieu},
  \bibinfo{author}{A.~Dudzik}, \bibinfo{author}{J.~Chung},
  \bibinfo{author}{D.~H. Choi}, \bibinfo{author}{R.~Powell},
  \bibinfo{author}{T.~Ewalds}, \bibinfo{author}{P.~Georgiev},
  \bibinfo{author}{J.~Oh}, \bibinfo{author}{D.~Horgan},
  \bibinfo{author}{M.~Kroiss}, \bibinfo{author}{I.~Danihelka},
  \bibinfo{author}{A.~Huang}, \bibinfo{author}{L.~Sifre},
  \bibinfo{author}{T.~Cai}, \bibinfo{author}{J.~P. Agapiou},
  \bibinfo{author}{M.~Jaderberg}, \bibinfo{author}{A.~S. Vezhnevets},
  \bibinfo{author}{R.~Leblond}, \bibinfo{author}{T.~Pohlen},
  \bibinfo{author}{V.~Dalibard}, \bibinfo{author}{D.~Budden},
  \bibinfo{author}{Y.~Sulsky}, \bibinfo{author}{J.~Molloy},
  \bibinfo{author}{T.~L. Paine}, \bibinfo{author}{C.~Gulcehre},
  \bibinfo{author}{Z.~Wang}, \bibinfo{author}{T.~Pfaff},
  \bibinfo{author}{Y.~Wu}, \bibinfo{author}{R.~Ring},
  \bibinfo{author}{D.~Yogatama}, \bibinfo{author}{D.~W{\"{u}}nsch},
  \bibinfo{author}{K.~McKinney}, \bibinfo{author}{O.~Smith},
  \bibinfo{author}{T.~Schaul}, \bibinfo{author}{T.~Lillicrap},
  \bibinfo{author}{K.~Kavukcuoglu}, \bibinfo{author}{D.~Hassabis},
  \bibinfo{author}{C.~Apps}, \bibinfo{author}{D.~Silver},
\newblock \bibinfo{title}{{Grandmaster level in StarCraft II using multi-agent
  reinforcement learning}},
\newblock \bibinfo{journal}{Nature} \bibinfo{volume}{575}
  (\bibinfo{year}{2019}) \bibinfo{pages}{350--354}.
\bibitem[{Wang and Hong(2020)}]{wang2020reinforcement}
\bibinfo{author}{Z.~Wang}, \bibinfo{author}{T.~Hong},
\newblock \bibinfo{title}{Reinforcement learning for building controls: The
  opportunities and challenges},
\newblock \bibinfo{journal}{Applied Energy} \bibinfo{volume}{269}
  (\bibinfo{year}{2020}) \bibinfo{pages}{115036}.
\bibitem[{Nweye et~al.(2022)Nweye, Liu, Stone, and Nagy}]{NWEYE2022100202}
\bibinfo{author}{K.~Nweye}, \bibinfo{author}{B.~Liu},
  \bibinfo{author}{P.~Stone}, \bibinfo{author}{Z.~Nagy},
\newblock \bibinfo{title}{Real-world challenges for multi-agent reinforcement
  learning in grid-interactive buildings},
\newblock \bibinfo{journal}{Energy and AI} \bibinfo{volume}{10}
  (\bibinfo{year}{2022}) \bibinfo{pages}{100202}.
\bibitem[{Fochesato et~al.(2022)Fochesato, Khayatian, Lima, and
  Nagy}]{10.1145/3563357.3564080}
\bibinfo{author}{M.~Fochesato}, \bibinfo{author}{F.~Khayatian},
  \bibinfo{author}{D.~F. Lima}, \bibinfo{author}{Z.~Nagy},
\newblock \bibinfo{title}{On the use of conditional timegan to enhance the
  robustness of a reinforcement learning agent in the building domain},
\newblock in: \bibinfo{booktitle}{Proceedings of the 9th ACM International
  Conference on Systems for Energy-Efficient Buildings, Cities, and
  Transportation}, BuildSys '22, \bibinfo{publisher}{Association for Computing
  Machinery}, \bibinfo{address}{New York, NY, USA}, \bibinfo{year}{2022}, p.
  \bibinfo{pages}{208–217}. \URLprefix
  \url{https://doi.org/10.1145/3563357.3564080}.
  \DOIprefix\doi{10.1145/3563357.3564080}.
\bibitem[{Pinto et~al.(2022)Pinto, Wang, Roy, Hong, and
  Capozzoli}]{pinto2022transfer}
\bibinfo{author}{G.~Pinto}, \bibinfo{author}{Z.~Wang},
  \bibinfo{author}{A.~Roy}, \bibinfo{author}{T.~Hong},
  \bibinfo{author}{A.~Capozzoli},
\newblock \bibinfo{title}{Transfer learning for smart buildings: A critical
  review of algorithms, applications, and future perspectives},
\newblock \bibinfo{journal}{Advances in Applied Energy}  (\bibinfo{year}{2022})
  \bibinfo{pages}{100084}.
\bibitem[{Vazquez-Canteli et~al.(2020)Vazquez-Canteli, Dey, Henze, and
  Nagy}]{citylearnArxiv}
\bibinfo{author}{J.~R. Vazquez-Canteli}, \bibinfo{author}{S.~Dey},
  \bibinfo{author}{G.~Henze}, \bibinfo{author}{Z.~Nagy},
\newblock \bibinfo{title}{{CityLearn: Standardizing Research in Multi-Agent
  Reinforcement Learning for Demand Response and Urban Energy Management}},
\newblock \bibinfo{journal}{arXiv}  (\bibinfo{year}{2020}).
\bibitem[{Blum et~al.(2021)Blum, Arroyo, Huang, Drgo{\v n}a, Jorissen, Walnum,
  Chen, Benne, Vrabie, Wetter, and Helsen}]{doi:10.1080/19401493.2021.1986574}
\bibinfo{author}{D.~Blum}, \bibinfo{author}{J.~Arroyo},
  \bibinfo{author}{S.~Huang}, \bibinfo{author}{J.~Drgo{\v n}a},
  \bibinfo{author}{F.~Jorissen}, \bibinfo{author}{H.~T. Walnum},
  \bibinfo{author}{Y.~Chen}, \bibinfo{author}{K.~Benne},
  \bibinfo{author}{D.~Vrabie}, \bibinfo{author}{M.~Wetter},
  \bibinfo{author}{L.~Helsen},
\newblock \bibinfo{title}{Building optimization testing framework (boptest) for
  simulation-based benchmarking of control strategies in buildings},
\newblock \bibinfo{journal}{Journal of Building Performance Simulation}
  \bibinfo{volume}{14} (\bibinfo{year}{2021}) \bibinfo{pages}{586--610}.
\bibitem[{Marzullo et~al.(2022)Marzullo, Dey, Long, Vilaplana, and
  Henze}]{Marzullo2022}
\bibinfo{author}{T.~Marzullo}, \bibinfo{author}{S.~Dey},
  \bibinfo{author}{N.~Long}, \bibinfo{author}{J.~L. Vilaplana},
  \bibinfo{author}{G.~Henze},
\newblock \bibinfo{title}{A high-fidelity building performance simulation test
  bed for the development and evaluation of advanced controls},
\newblock \bibinfo{journal}{Journal of Building Performance Simulation}
  \bibinfo{volume}{15} (\bibinfo{year}{2022}) \bibinfo{pages}{379--397}.
\bibitem[{Scharnhorst et~al.(2021)Scharnhorst, Schubnel, Fern{\'a}ndez~Bandera,
  Salom, Taddeo, Boegli, Gorecki, Stauffer, Peppas, and
  Politi}]{scharnhorst2021energym}
\bibinfo{author}{P.~Scharnhorst}, \bibinfo{author}{B.~Schubnel},
  \bibinfo{author}{C.~Fern{\'a}ndez~Bandera}, \bibinfo{author}{J.~Salom},
  \bibinfo{author}{P.~Taddeo}, \bibinfo{author}{M.~Boegli},
  \bibinfo{author}{T.~Gorecki}, \bibinfo{author}{Y.~Stauffer},
  \bibinfo{author}{A.~Peppas}, \bibinfo{author}{C.~Politi},
\newblock \bibinfo{title}{Energym: A building model library for controller
  benchmarking},
\newblock \bibinfo{journal}{Applied Sciences} \bibinfo{volume}{11}
  (\bibinfo{year}{2021}) \bibinfo{pages}{3518}.
\bibitem[{Electric Power Research Institute(2017)}]{epriFontana2017}
Electric Power Research Institute, \bibinfo{title}{Grid integration of zero net
  energy communities}, \bibinfo{year}{2017}. \URLprefix
  \url{https://www.calmac.org/publications/CSIRDD_Sol4_EPRI_Grid-Integration-of-ZNE-Communities_FinalRpt_2017-01-27.pdf}.
\bibitem[{Narayanamurthy et~al.(2016)Narayanamurthy, Handa, Tumilowicz, Herro,
  and Shah}]{narayanamurthy2016grid}
\bibinfo{author}{R.~Narayanamurthy}, \bibinfo{author}{R.~Handa},
  \bibinfo{author}{N.~Tumilowicz}, \bibinfo{author}{C.~Herro},
  \bibinfo{author}{S.~Shah},
\newblock \bibinfo{title}{Grid integration of zero net energy communities},
\newblock \bibinfo{journal}{ACEEE Summer Study Energy Effic. Build}
  (\bibinfo{year}{2016}).
\bibitem[{NeurIPS 2022 --The CityLearn Challenge
  2022(2022)}]{AICrowdCityLearnChallenge2022}
NeurIPS 2022 --The CityLearn Challenge 2022, \bibinfo{title}{November 2022
  monthly energy review}, \bibinfo{year}{2022}. \URLprefix
  \url{https://www.aicrowd.com/challenges/neurips-2022-citylearn-challenge}.
\bibitem[{V{\'{a}}zquez-Canteli et~al.(2020)V{\'{a}}zquez-Canteli, Dey, Henze,
  and Nagy}]{Vazquez-Canteli2020a}
\bibinfo{author}{J.~R. V{\'{a}}zquez-Canteli}, \bibinfo{author}{S.~Dey},
  \bibinfo{author}{G.~Henze}, \bibinfo{author}{Z.~Nagy},
\newblock \bibinfo{title}{{The CityLearn Challenge 2020}},
\newblock in: \bibinfo{booktitle}{Proceedings of the 7th ACM International
  Conference on Systems for Energy-Efficient Buildings, Cities, and
  Transportation}, \bibinfo{publisher}{ACM}, \bibinfo{address}{New York, NY,
  USA}, \bibinfo{year}{2020}, pp. \bibinfo{pages}{320--321}. \URLprefix
  \url{https://dl.acm.org/doi/10.1145/3408308.3431122}.
  \DOIprefix\doi{10.1145/3408308.3431122}.
\bibitem[{Nagy et~al.(2021)Nagy, V{\'{a}}zquez-Canteli, Dey, and
  Henze}]{Nagy2021}
\bibinfo{author}{Z.~Nagy}, \bibinfo{author}{J.~R. V{\'{a}}zquez-Canteli},
  \bibinfo{author}{S.~Dey}, \bibinfo{author}{G.~Henze},
\newblock \bibinfo{title}{{The citylearn challenge 2021}},
\newblock in: \bibinfo{booktitle}{Proceedings of the 8th ACM International
  Conference on Systems for Energy-Efficient Buildings, Cities, and
  Transportation}, \bibinfo{publisher}{ACM}, \bibinfo{address}{New York, NY,
  USA}, \bibinfo{year}{2021}, pp. \bibinfo{pages}{218--219}. \URLprefix
  \url{https://dl.acm.org/doi/10.1145/3486611.3492226}.
  \DOIprefix\doi{10.1145/3486611.3492226}.
\bibitem[{Southern California Edison Time-Of-Use Residential Rate
  Plans(2022)}]{sCEDTOU}
Southern California Edison Time-Of-Use Residential Rate Plans,
  \bibinfo{title}{Time-of-use (tou) rate plans}, \bibinfo{year}{2022}.
  \URLprefix
  \url{https://www.sce.com/residential/rates/Time-Of-Use-Residential-Rate-Plans}.
\bibitem[{Deltetto et~al.(2021)Deltetto, Coraci, Pinto, Piscitelli, and
  Capozzoli}]{Deltetto2021}
\bibinfo{author}{D.~Deltetto}, \bibinfo{author}{D.~Coraci},
  \bibinfo{author}{G.~Pinto}, \bibinfo{author}{M.~S. Piscitelli},
  \bibinfo{author}{A.~Capozzoli},
\newblock \bibinfo{title}{{Exploring the potentialities of deep reinforcement
  learning for incentive-based demand response in a cluster of small commercial
  buildings}},
\newblock \bibinfo{journal}{Energies} \bibinfo{volume}{14}
  (\bibinfo{year}{2021}).
\bibitem[{Glatt et~al.(2021)Glatt, Silva, Soper, Dawson, Rusu, and
  Goldhahn}]{Glatt2021}
\bibinfo{author}{R.~Glatt}, \bibinfo{author}{F.~L.~d. Silva},
  \bibinfo{author}{B.~Soper}, \bibinfo{author}{W.~A. Dawson},
  \bibinfo{author}{E.~Rusu}, \bibinfo{author}{R.~A. Goldhahn},
\newblock \bibinfo{title}{{Collaborative energy demand response with
  decentralized actor and centralized critic}},
\newblock in: \bibinfo{booktitle}{Proceedings of the 8th ACM International
  Conference on Systems for Energy-Efficient Buildings, Cities, and
  Transportation}, \bibinfo{publisher}{ACM}, \bibinfo{address}{New York, NY,
  USA}, \bibinfo{year}{2021}, pp. \bibinfo{pages}{333--337}. \URLprefix
  \url{https://dl.acm.org/doi/10.1145/3486611.3488732}.
  \DOIprefix\doi{10.1145/3486611.3488732}.
\bibitem[{Pinto et~al.(2021)Pinto, Piscitelli, V{\'{a}}zquez-Canteli, Nagy, and
  Capozzoli}]{Pinto2021}
\bibinfo{author}{G.~Pinto}, \bibinfo{author}{M.~S. Piscitelli},
  \bibinfo{author}{J.~R. V{\'{a}}zquez-Canteli}, \bibinfo{author}{Z.~Nagy},
  \bibinfo{author}{A.~Capozzoli},
\newblock \bibinfo{title}{{Coordinated energy management for a cluster of
  buildings through deep reinforcement learning}},
\newblock \bibinfo{journal}{Energy} \bibinfo{volume}{229}
  (\bibinfo{year}{2021}).
\bibitem[{Kathirgamanathan et~al.(2020)Kathirgamanathan, Twardowski, Mangina,
  and Finn}]{Kathirgamanathan2020}
\bibinfo{author}{A.~Kathirgamanathan}, \bibinfo{author}{K.~Twardowski},
  \bibinfo{author}{E.~Mangina}, \bibinfo{author}{D.~P. Finn},
\newblock \bibinfo{title}{{A Centralised Soft Actor Critic Deep Reinforcement
  Learning Approach to District Demand Side Management through CityLearn}},
\newblock in: \bibinfo{booktitle}{Proceedings of the 1st International Workshop
  on Reinforcement Learning for Energy Management in Buildings {\&} Cities},
  \bibinfo{publisher}{ACM}, \bibinfo{address}{New York, NY, USA},
  \bibinfo{year}{2020}, pp. \bibinfo{pages}{11--14}. \URLprefix
  \url{https://dl.acm.org/doi/10.1145/3427773.3427869}.
  \DOIprefix\doi{10.1145/3427773.3427869}.
\bibitem[{Pinto et~al.(2021)Pinto, Deltetto, and Capozzoli}]{pinto2021data}
\bibinfo{author}{G.~Pinto}, \bibinfo{author}{D.~Deltetto},
  \bibinfo{author}{A.~Capozzoli},
\newblock \bibinfo{title}{Data-driven district energy management with surrogate
  models and deep reinforcement learning},
\newblock \bibinfo{journal}{Applied Energy} \bibinfo{volume}{304}
  (\bibinfo{year}{2021}) \bibinfo{pages}{117642}.
\bibitem[{Dhamankar et~al.(2020)Dhamankar, Vazquez-Canteli, and
  Nagy}]{Dhamankar2020}
\bibinfo{author}{G.~Dhamankar}, \bibinfo{author}{J.~R. Vazquez-Canteli},
  \bibinfo{author}{Z.~Nagy},
\newblock \bibinfo{title}{{Benchmarking Multi-Agent Deep Reinforcement Learning
  Algorithms on a Building Energy Demand Coordination Task}},
\newblock \bibinfo{journal}{RLEM 2020 - Proceedings of the 1st International
  Workshop on Reinforcement Learning for Energy Management in Buildings and
  Cities}  (\bibinfo{year}{2020}) \bibinfo{pages}{15--19}.
\bibitem[{Qin et~al.(2021)Qin, Gao, Zhang, Xu, Huang, Li, Zhang, and
  Yu}]{Qin2021}
\bibinfo{author}{R.~Qin}, \bibinfo{author}{S.~Gao}, \bibinfo{author}{X.~Zhang},
  \bibinfo{author}{Z.~Xu}, \bibinfo{author}{S.~Huang}, \bibinfo{author}{Z.~Li},
  \bibinfo{author}{W.~Zhang}, \bibinfo{author}{Y.~Yu},
\newblock \bibinfo{title}{{NeoRL: A Near Real-World Benchmark for Offline
  Reinforcement Learning}}  (\bibinfo{year}{2021}).
\bibitem[{Pinto et~al.(2022)Pinto, Kathirgamanathan, Mangina, Finn, and
  Capozzoli}]{pinto2022enhancing}
\bibinfo{author}{G.~Pinto}, \bibinfo{author}{A.~Kathirgamanathan},
  \bibinfo{author}{E.~Mangina}, \bibinfo{author}{D.~P. Finn},
  \bibinfo{author}{A.~Capozzoli},
\newblock \bibinfo{title}{Enhancing energy management in grid-interactive
  buildings: A comparison among cooperative and coordinated architectures},
\newblock \bibinfo{journal}{Applied Energy} \bibinfo{volume}{310}
  (\bibinfo{year}{2022}) \bibinfo{pages}{118497}.
\bibitem[{Pigott et~al.(2022)Pigott, Crozier, Baker, and
  Nagy}]{PIGOTT2022108521}
\bibinfo{author}{A.~Pigott}, \bibinfo{author}{C.~Crozier},
  \bibinfo{author}{K.~Baker}, \bibinfo{author}{Z.~Nagy},
\newblock \bibinfo{title}{Gridlearn: Multiagent reinforcement learning for
  grid-aware building energy management},
\newblock \bibinfo{journal}{Electric Power Systems Research}
  \bibinfo{volume}{213} (\bibinfo{year}{2022}) \bibinfo{pages}{108521}.
\bibitem[{Haarnoja et~al.(2018{\natexlab{a}})Haarnoja, Zhou, Abbeel, and
  Levine}]{Haarnoja2018}
\bibinfo{author}{T.~Haarnoja}, \bibinfo{author}{A.~Zhou},
  \bibinfo{author}{P.~Abbeel}, \bibinfo{author}{S.~Levine},
\newblock \bibinfo{title}{{Soft Actor-Critic: Off-Policy Maximum Entropy Deep
  Reinforcement Learning with a Stochastic Actor}},
\newblock in: \bibinfo{booktitle}{ICML}, \bibinfo{year}{2018}{\natexlab{a}}.
  \URLprefix \url{http://arxiv.org/abs/1801.01290}.
\bibitem[{Haarnoja et~al.(2018{\natexlab{b}})Haarnoja, Zhou, Hartikainen,
  Tucker, Ha, Tan, Kumar, Zhu, Gupta, Abbeel, and Levine}]{Haarnoja2018SoftAA}
\bibinfo{author}{T.~Haarnoja}, \bibinfo{author}{A.~Zhou},
  \bibinfo{author}{K.~Hartikainen}, \bibinfo{author}{G.~Tucker},
  \bibinfo{author}{S.~Ha}, \bibinfo{author}{J.~Tan},
  \bibinfo{author}{V.~Kumar}, \bibinfo{author}{H.~Zhu},
  \bibinfo{author}{A.~Gupta}, \bibinfo{author}{P.~Abbeel},
  \bibinfo{author}{S.~Levine},
\newblock \bibinfo{title}{Soft actor-critic algorithms and applications},
\newblock \bibinfo{journal}{ArXiv} \bibinfo{volume}{abs/1812.05905}
  (\bibinfo{year}{2018}{\natexlab{b}}).
\bibitem[{Vazquez-Canteli et~al.(2020)Vazquez-Canteli, Henze, and
  Nagy}]{Vazquez-Canteli2020}
\bibinfo{author}{J.~R. Vazquez-Canteli}, \bibinfo{author}{G.~Henze},
  \bibinfo{author}{Z.~Nagy},
\newblock \bibinfo{title}{{MARLISA: Multi-Agent Reinforcement Learning with
  Iterative Sequential Action Selection for Load Shaping of Grid-Interactive
  Connected Buildings}},
\newblock \bibinfo{journal}{BuildSys 2020 - Proceedings of the 7th ACM
  International Conference on Systems for Energy-Efficient Buildings, Cities,
  and Transportation}  (\bibinfo{year}{2020}) \bibinfo{pages}{170--179}.
\bibitem[{Park et~al.(2019)Park, Ouf, Gunay, Peng, O'Brien, Kj{\ae}rgaard, and
  Nagy}]{park2019critical}
\bibinfo{author}{J.~Y. Park}, \bibinfo{author}{M.~M. Ouf},
  \bibinfo{author}{B.~Gunay}, \bibinfo{author}{Y.~Peng},
  \bibinfo{author}{W.~O'Brien}, \bibinfo{author}{M.~B. Kj{\ae}rgaard},
  \bibinfo{author}{Z.~Nagy},
\newblock \bibinfo{title}{A critical review of field implementations of
  occupant-centric building controls},
\newblock \bibinfo{journal}{Building and Environment} \bibinfo{volume}{165}
  (\bibinfo{year}{2019}) \bibinfo{pages}{106351}.
\bibitem[{Quintana et~al.(2022)Quintana, Tartarini, Schiavon, and
  Miller}]{Quintana2022}
\bibinfo{author}{M.~Quintana}, \bibinfo{author}{F.~Tartarini},
  \bibinfo{author}{S.~Schiavon}, \bibinfo{author}{C.~Miller},
\newblock \bibinfo{title}{{ComfortLearn : Enabling agent-based occupant-centric
  building controls}}  (\bibinfo{year}{2022}) \bibinfo{pages}{475--478}.
\bibitem[{Dulac-Arnold et~al.(2021)Dulac-Arnold, Levine, Mankowitz, Li,
  Paduraru, Gowal, and Hester}]{dulac2021challenges}
\bibinfo{author}{G.~Dulac-Arnold}, \bibinfo{author}{N.~Levine},
  \bibinfo{author}{D.~J. Mankowitz}, \bibinfo{author}{J.~Li},
  \bibinfo{author}{C.~Paduraru}, \bibinfo{author}{S.~Gowal},
  \bibinfo{author}{T.~Hester},
\newblock \bibinfo{title}{Challenges of real-world reinforcement learning:
  definitions, benchmarks and analysis},
\newblock \bibinfo{journal}{Machine Learning} \bibinfo{volume}{110}
  (\bibinfo{year}{2021}) \bibinfo{pages}{2419--2468}.
\bibitem[{Xilei et~al.(2022)Xilei, Cheng, and Chong}]{Xilei2022}
\bibinfo{author}{D.~Xilei}, \bibinfo{author}{S.~Cheng},
  \bibinfo{author}{A.~Chong},
\newblock \bibinfo{title}{Deciphering optimal mixed-mode ventilation in the
  tropics using reinforcement learning with explainable artificial
  intelligence},
\newblock \bibinfo{journal}{Energy and Buildings} \bibinfo{volume}{278}
  (\bibinfo{year}{2022}) \bibinfo{pages}{112629}.

\end{thebibliography}
\balance

\clearpage
\onecolumn
\appendix
\renewcommand\thefigure{\thesection.\arabic{figure}}

\section{Figures} \label{sec:appendix-figures}
\setcounter{figure}{0}
\begin{figure}[!htb]
    \centering
    \includegraphics[width=\textwidth]{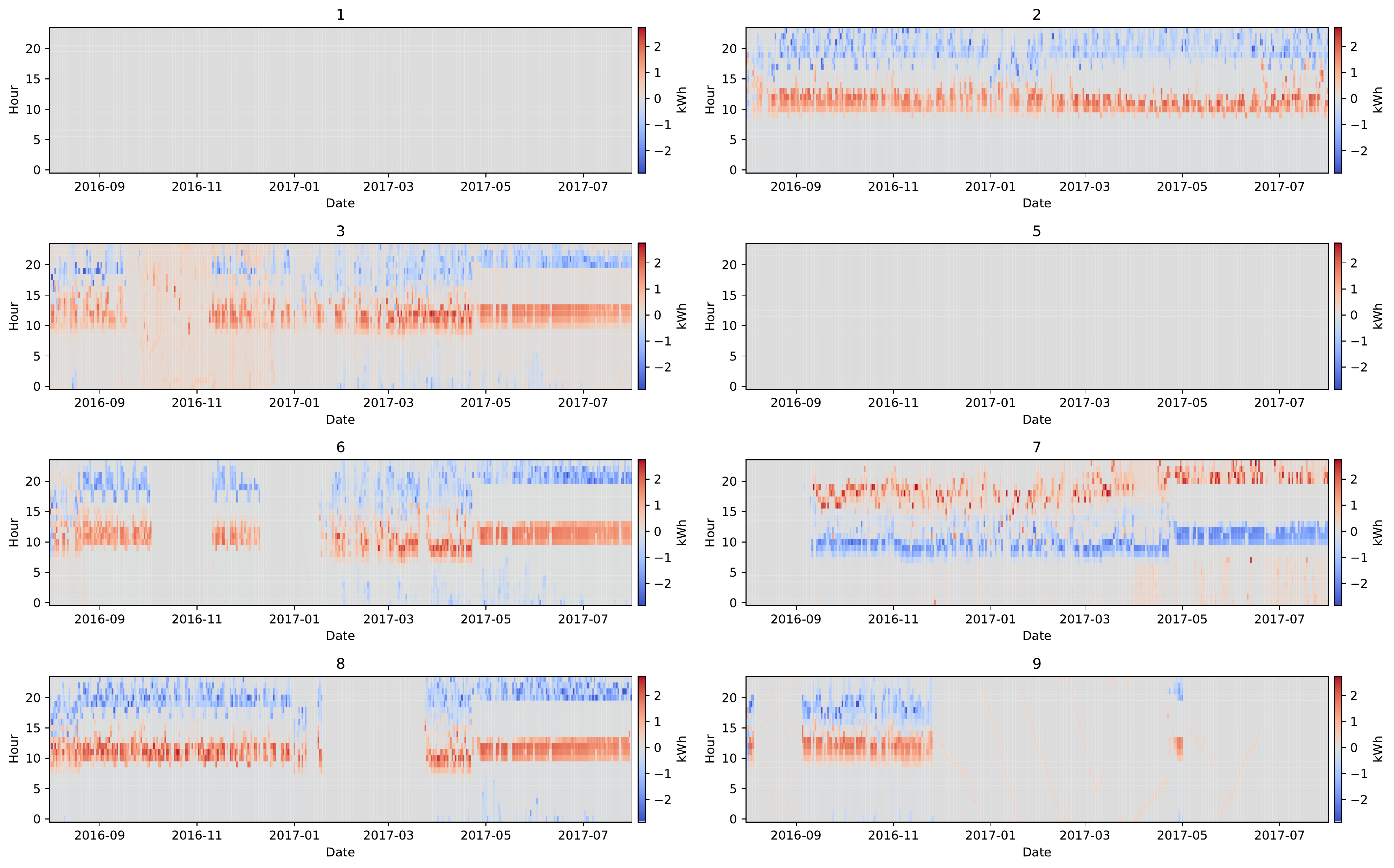}
    \caption{Calculated battery electricity consumption from smart meter-measured electric power demand for buildings in Sierra Crest Zero Net Energy community that have battery resource. Batteries in buildings 1 and 5 are not operational.}
    \label{fig:measured_battery_electricity_consumption}
\end{figure}

\begin{figure}[!htb]
    \centering
    \includegraphics[width=0.65\textwidth]{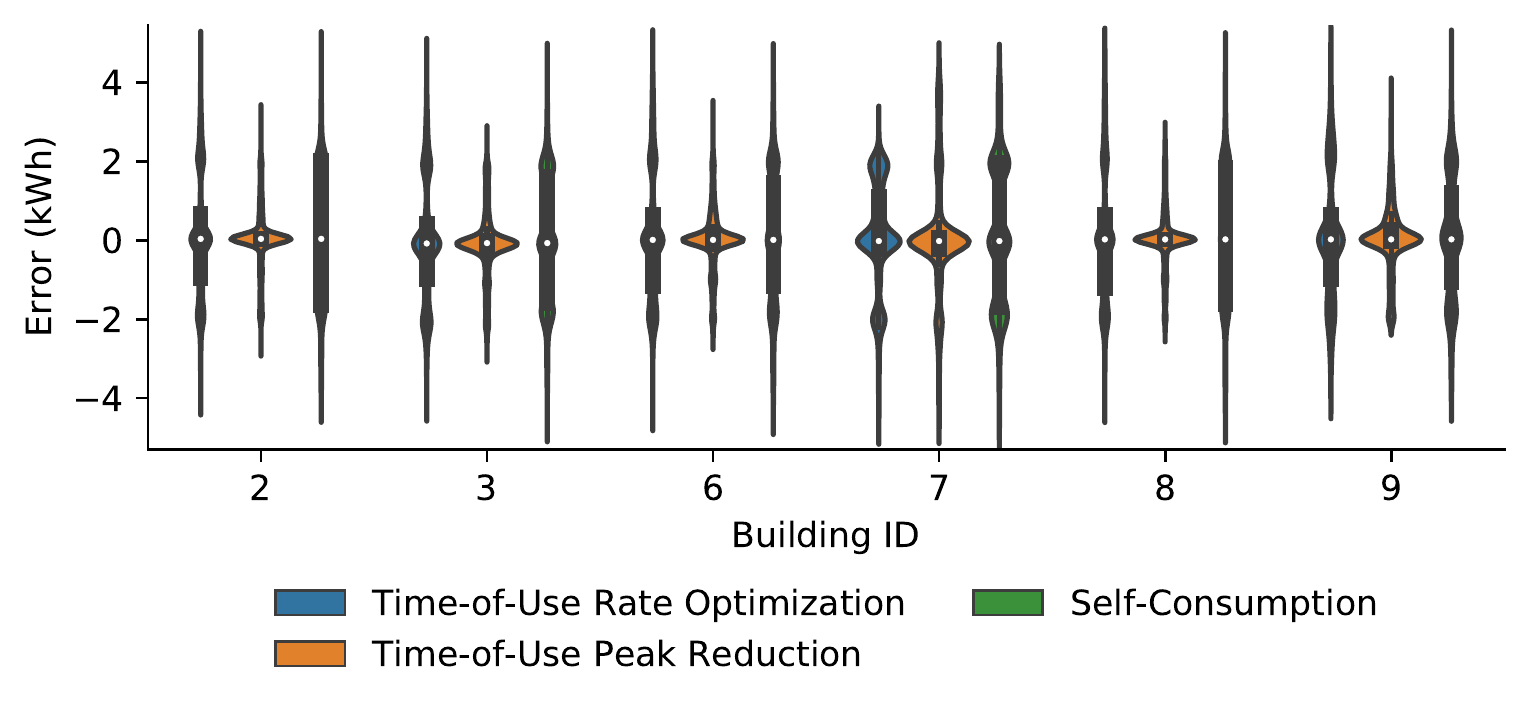}
    \caption{Difference between simulated and measured battery electricity consumption in digital twin and real-world communities respectively for three battery control strategies. While all three battery control strategies have their median error at $\approx$ 0.0 kWh, the \textit{Time-of-Use Peak Reduction} strategy has the least variance compared to the other two strategies for all buildings.}
    \label{fig:rbc_validation}
\end{figure}

\begin{figure}[!htb]
    \centering
    \begin{subfigure}[]{.35\textwidth}
        \centering
        \includegraphics[width=\textwidth]{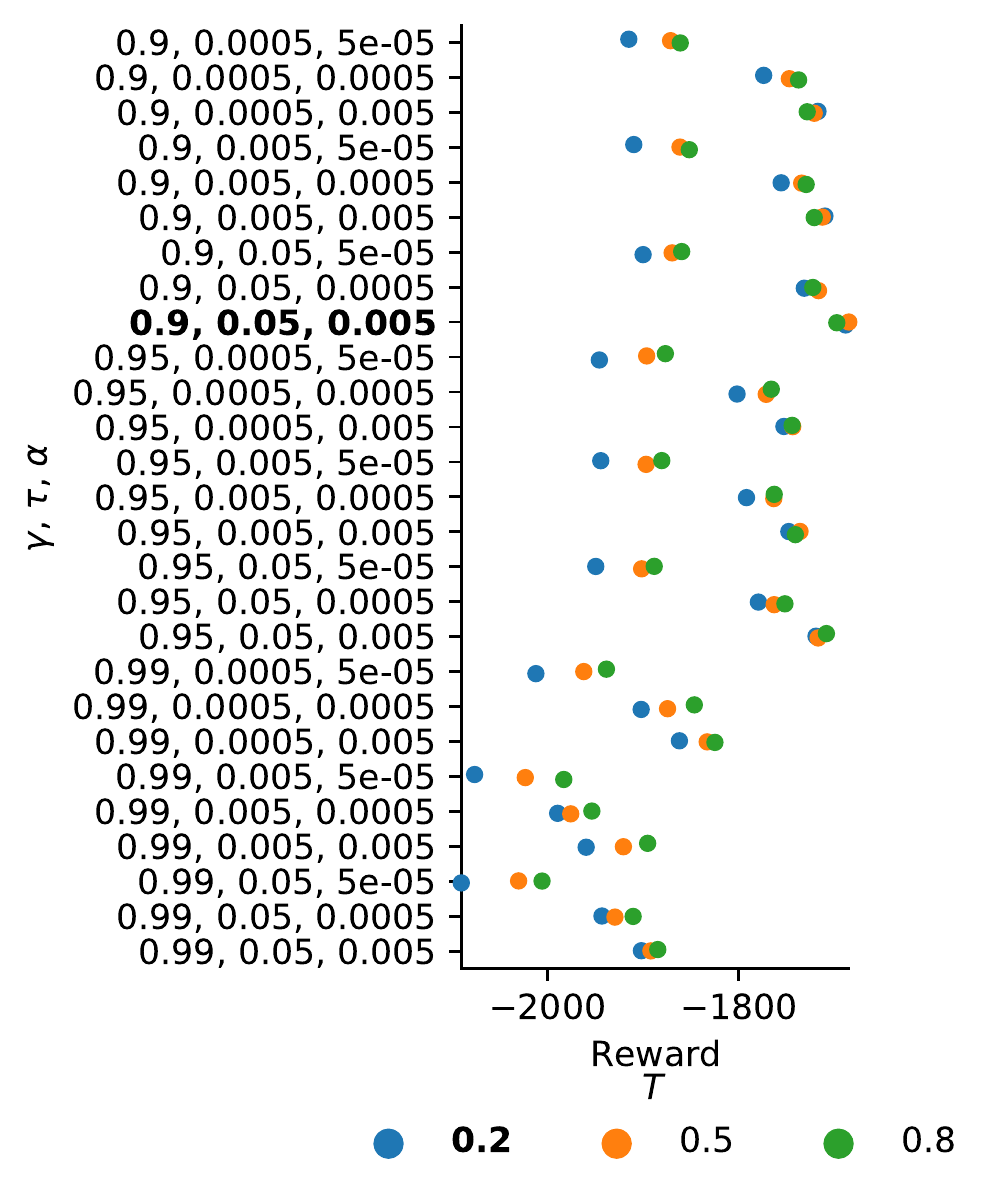}
        \caption{$T$ with respect to $\gamma$, $\tau$ and $\alpha$. When $\gamma \ge 0.95$, $T=0.8$ maximizes the average cumulative reward while at $\gamma = 0.90$, either $T=0.5$ or $T=0.8$ maximize the average cumulative reward.}
        \label{fig:last_episode_hyperparameter_combination_constant_alpha_reward}
    \end{subfigure}
    \begin{subfigure}[]{.35\textwidth}
        \centering
        \includegraphics[width=\textwidth]{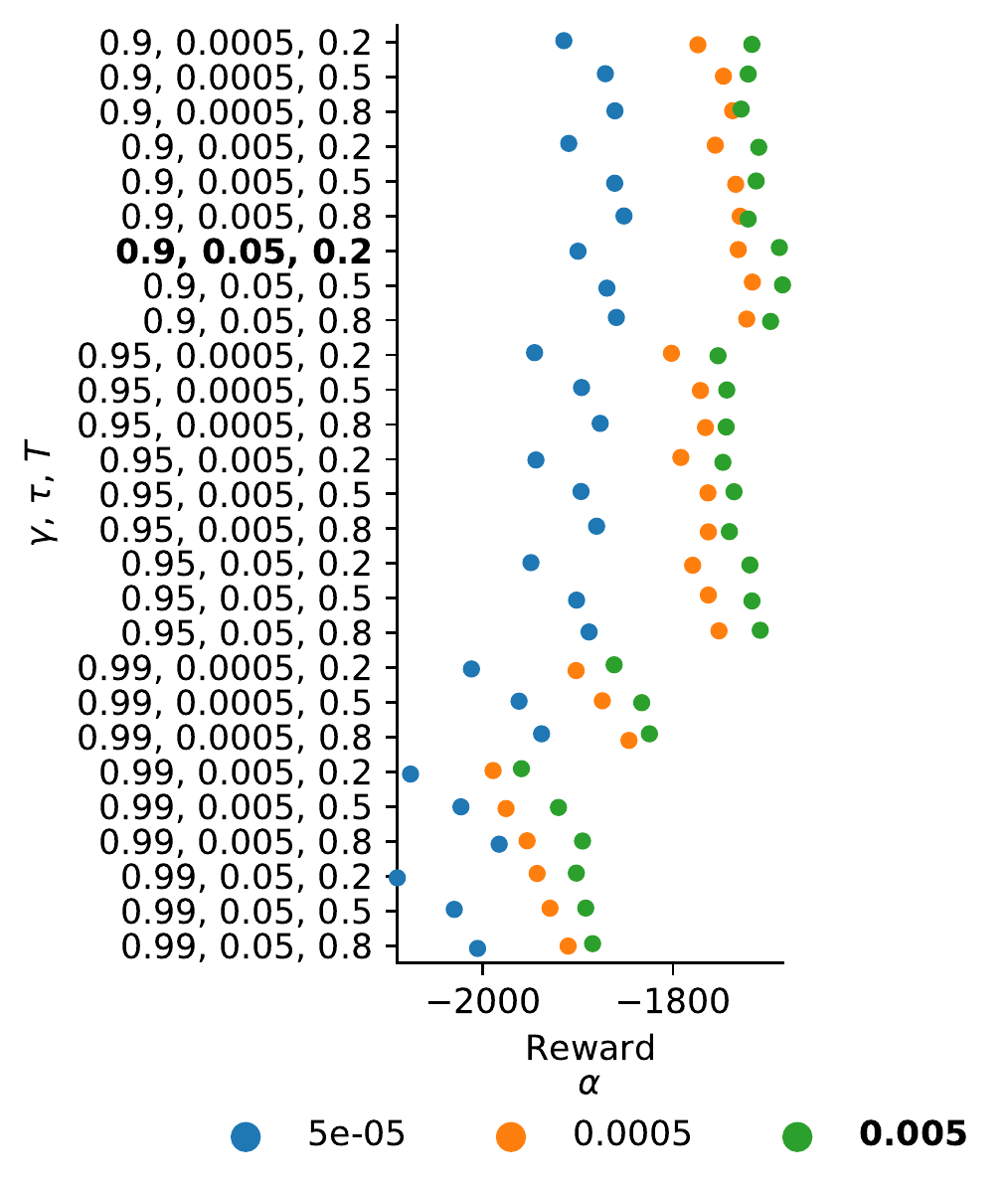}
        \caption{$\alpha$ with respect to $\gamma$, $\tau$ and $T$. For all combinations of $\gamma$, $\tau$ and $T$, the average cumulative reward is maximized as the learning rate, $\alpha$, increases.}
        \label{fig:last_episode_hyperparameter_combination_constant_lr_reward}
    \end{subfigure}
    \begin{subfigure}[]{.35\textwidth}
        \centering
        \includegraphics[width=\textwidth]{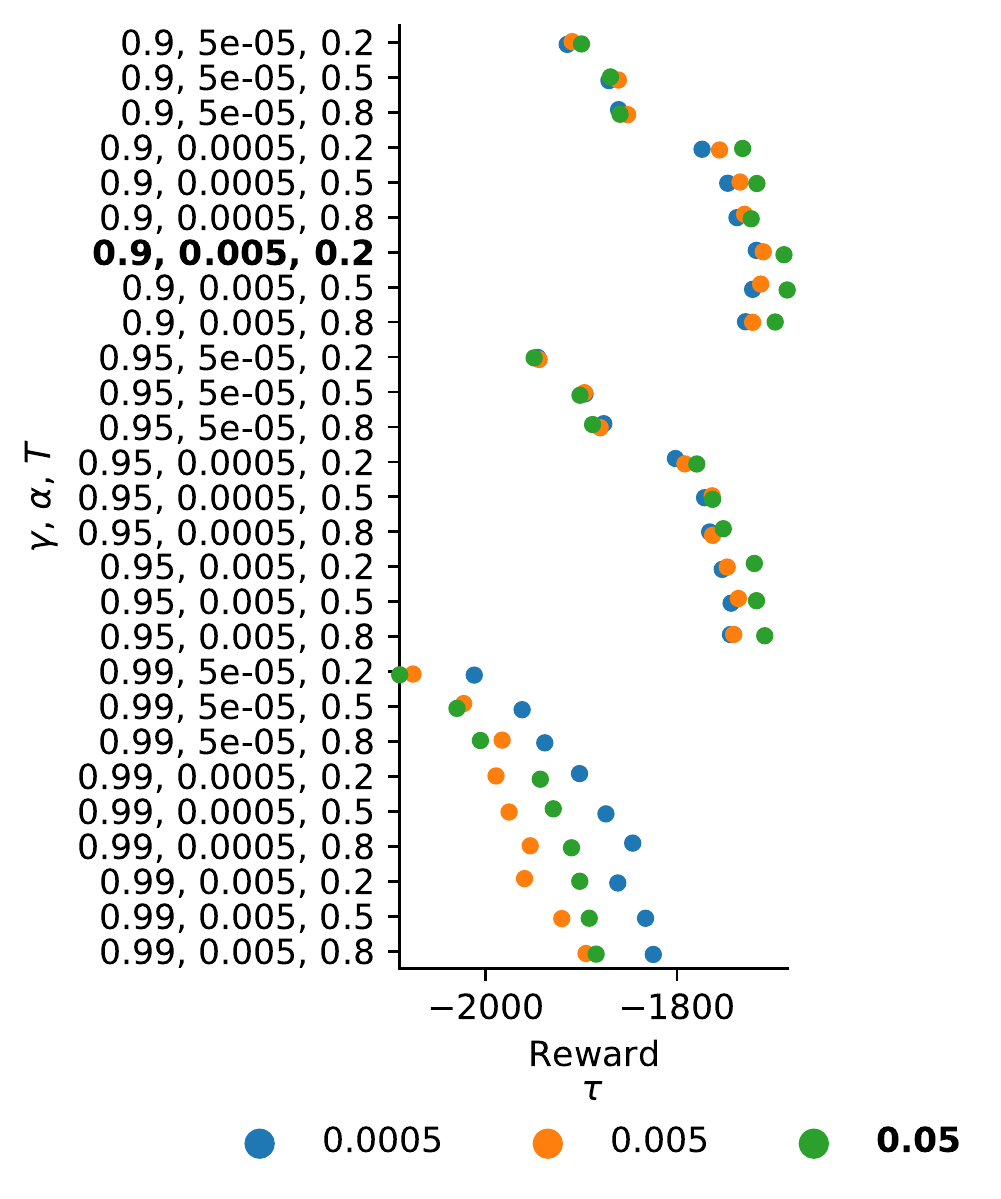}
        \caption{$\tau$ with respect to $\gamma$, $\alpha$ and $T$. Decreasing the decay rate, $\tau$ when $\gamma=0.99$ increases the average cumulative reward whereas, at lower values of $\gamma$, larger $\tau$ yields larger average cumulative reward.}
        \label{fig:last_episode_hyperparameter_combination_constant_tau_reward}
    \end{subfigure}
    \begin{subfigure}[]{.35\textwidth}
        \centering
        \includegraphics[width=\textwidth]{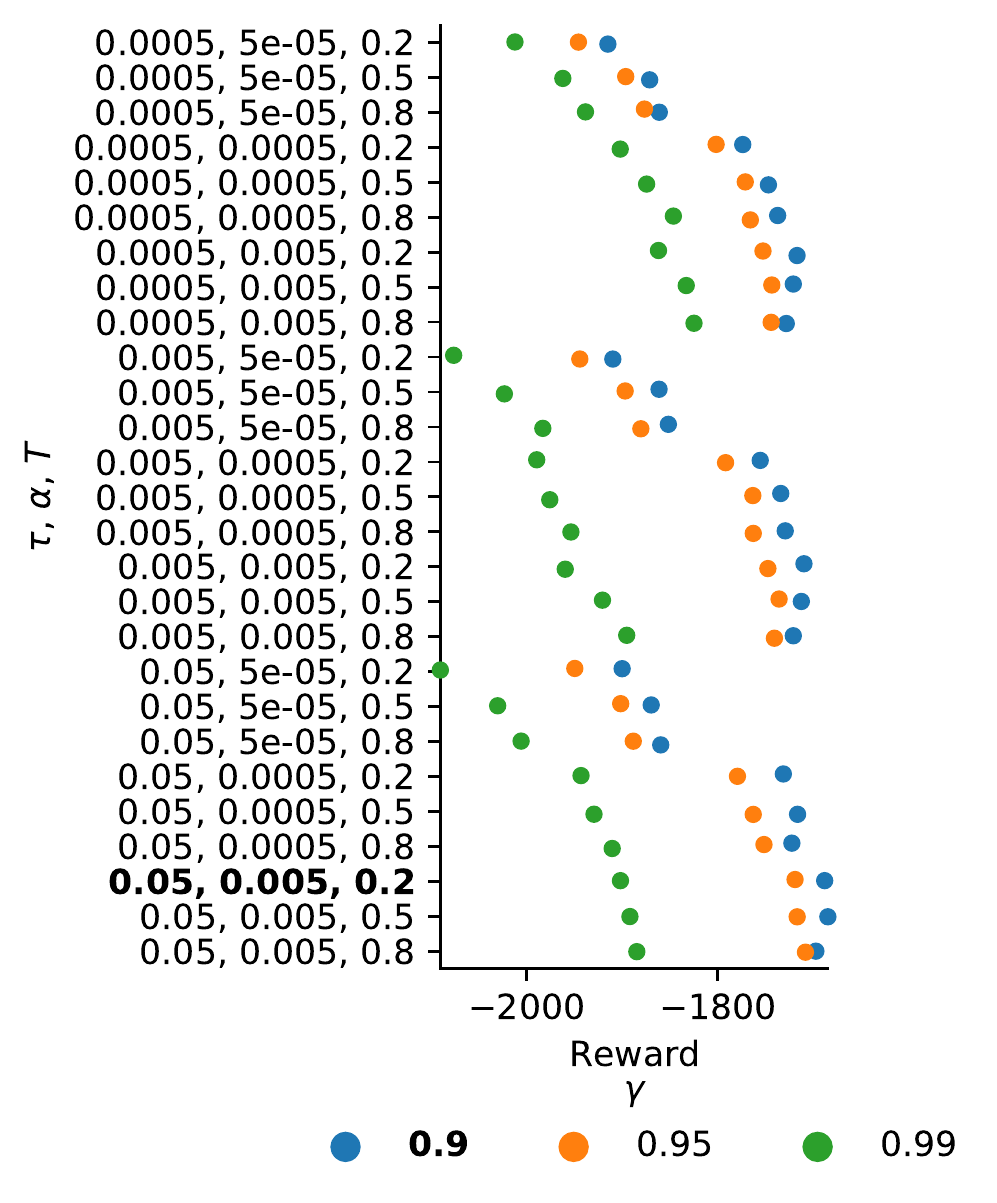}
        \caption{$\gamma$ with respect to $\tau$, $\alpha$ and $T$. Decreasing the discount factor, $\gamma$ brings about larger average cumulative reward for all combinations of $\tau$, $\alpha$ and $T$.}
        \label{fig:last_episode_hyperparameter_combination_constant_discount_reward}
    \end{subfigure}
    \caption{Average cumulative reward in the final training episode for different combinations RL agent hyperparameters: decay rate ($\tau$), discount factor ($\gamma$), learning rate ($\alpha$) and temperature ($T$). The hyperparameter value that is the mode of values that maximize the buildings' reward has a bold font and is selected as best-performing.}
    \label{fig:last_episode_hyperparameter_combination_reward}
\end{figure}

\begin{figure}[!htb]
    \centering
    \includegraphics[width=\textwidth]{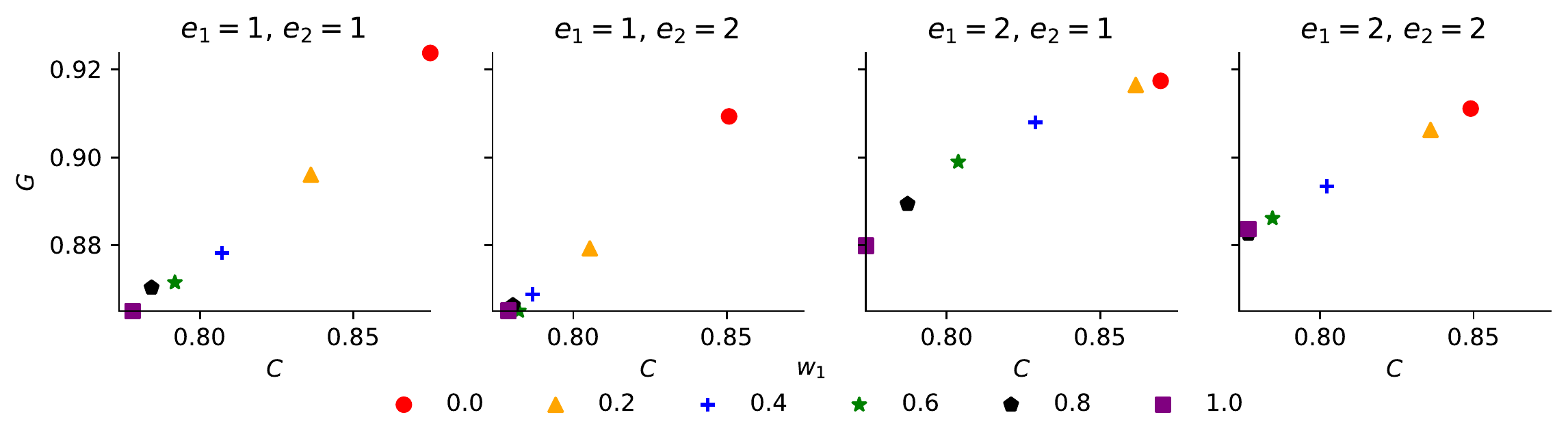}
    \caption{Distribution of the district-level electricity price, $C$, and carbon emissions, $G$, KPIs for different combinations of $e_1$, $e_2$, $w_1$ and $w_2$ ($1.0 - w_1$) in the last training episode of the reward parameter grid search. $C$ and $G$ are directly correlated and for each combination of $e_1$ and $e_2$, $w_1$ is indirectly proportional to both electricity price and carbon emissions score. Hence, increasing the weight on the electricity price yields better results. The lowest electricity price, $C$ score, $C=0.774$, is achieved when the parameters are ($e_1=2$, $e_2=1$, $w_1=1.0$, $w_2=0.0$). The parameters ($e_1=1$, $e_2=1$, $w_1=1.0$, $w_2=0.0$) minimize both $G$ and the average of $C$ and $G$ to 0.865 and 0.821 respectively while $C$ is slightly higher at 0.778. The aforementioned parameters attribute all weight to the electricity price component in the reward function.}
    \label{fig:reward_design_last_episode_district_price_and_emission_scores}
\end{figure}




\end{document}